# Sample Size in Natural Language Processing within Healthcare Research

Jaya Chaturvedi[1*], Diana Shamsutdinova[1], Felix Zimmer[1,3], Sumithra Velupillai[1], Daniel Stahl[1], Robert Stewart[1,2], Angus Roberts[1]
[1]Institute of Psychiatry, Psychology and Neurosciences, King's College London, London, United Kingdom
[2]South London and Maudsley NHS Foundation Trust, London, United Kingdom
[3]Psychological Institute of the University of Zurich, Zurich, Switzerland
*Corresponding author

## Abstract

### Objective

Sample size calculation is an essential step in most data-based disciplines. Large enough samples ensure representativeness of the population and determine the precision of estimates. This is true for most quantitative studies, including those that employ machine learning methods, such as natural language processing, where free-text is used to generate predictions and classify instances of text. Within the healthcare domain, the lack of sufficient corpora of previously collected data can be a limiting factor when determining sample sizes for new studies. This paper tries to address the issue by making recommendations on sample sizes for text classification tasks in the healthcare domain.

### Materials and Methods

Models trained on the MIMIC-III database of critical care records from Beth Israel Deaconess Medical Center were used to classify documents as having or not having Unspecified Essential Hypertension, the most common diagnosis code in the database. Simulations were performed using various classifiers on different sample sizes and class proportions. This was repeated for a comparatively less common diagnosis code within the database of diabetes mellitus without mention of complication.

### Results

A table listing the expected performances for different classifiers under varying conditions of sample size and class proportion is presented. Smaller sample sizes resulted in better results when using a K-nearest neighbours classifier, whereas larger sample sizes provided better results with support vector machines and BERT models. Overall, a sample size larger than 1000 was sufficient to provide decent performance metrics.

## Conclusion

The simulations conducted within this study provide guidelines that can be used as recommendations for selecting appropriate sample sizes and class proportions, and for predicting expected performance, when building classifiers for textual healthcare data. The methodology used here can be modified for sample size estimates calculations with other datasets.

## Keywords

sample size, natural language processing, machine learning

## Graphical Abstract

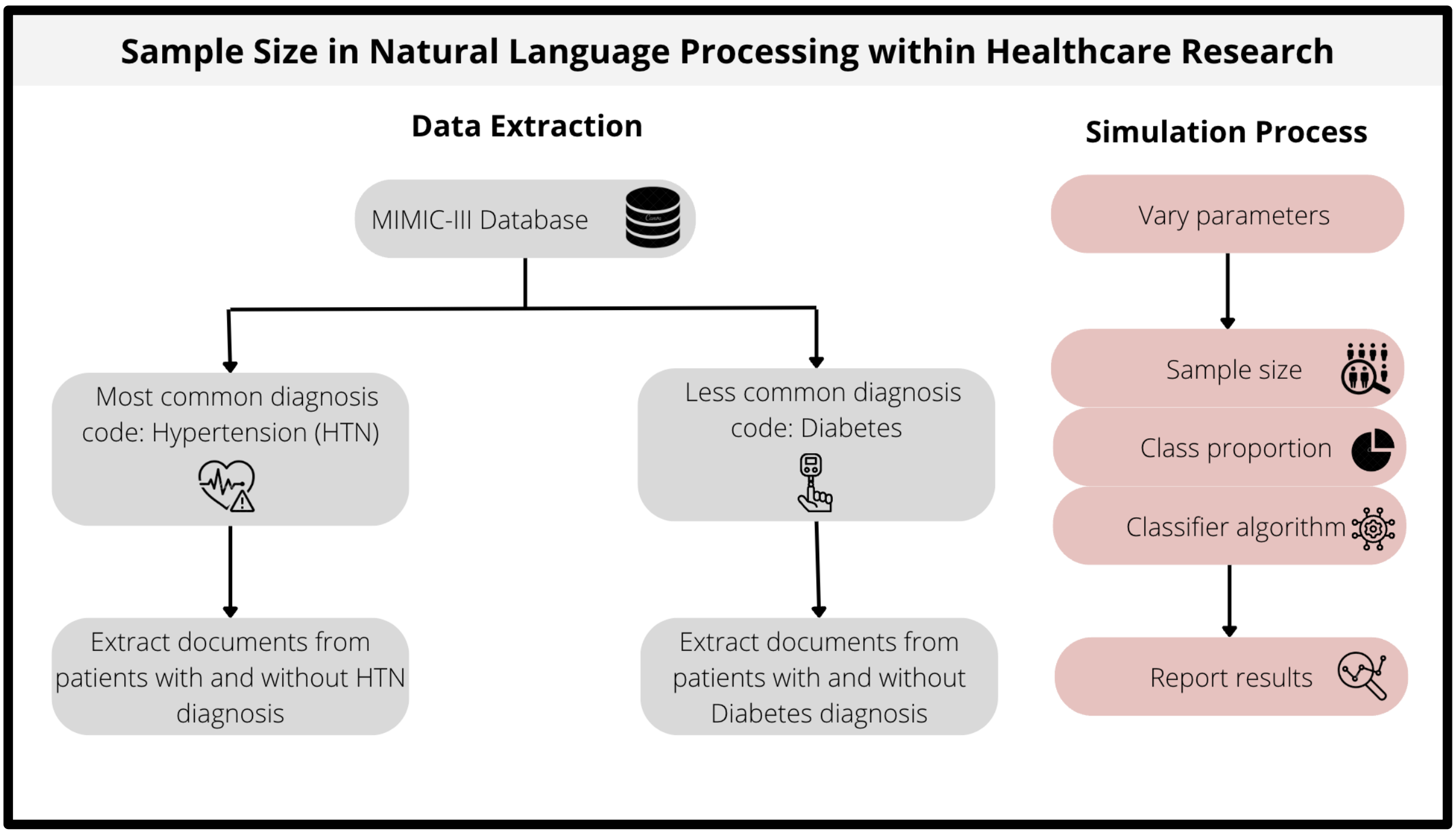

# 1. Introduction

A sample is a subset of a larger population. The aim of a good sample is for it to be representative of the larger population and provide results approximating what might be found when using the entire population [1]. Knowing whether an appropriate sample size was used is crucial to determine the value of a research project as it gives an insight into whether appropriate considerations were made to ensure the project is ethical and methodologically sound [2].

As with other types of research, appropriate sample size is essential for quantitative research, especially for the generalisability and reproducibility of the findings [3]. Most epidemiological studies focus on relationships between some exposure variables and disease outcomes [4]. In these instances, it is of importance to use a sample that is truly representative of the population of interest, in order to prevent inaccurate deductions from any statistical analyses conducted on these samples. While there are multiple factors that can result in unrepresentative samples, insufficient sample size is a particularly dangerous one. Although random samples are on average unbiased, a small sample will often fail to accurately represent the underlying population, leading to inaccurate observations regarding the relationship between predictors and outcomes. Along with sample size, it is important to use appropriate methods for ascertaining such samples, such as random sampling, in order to avoid magnifying unrepresentativeness within large samples [5]. Despite this being an important step in quantitative research, a study found that 60% of publications on such research do not provide details on sampling methods and approaches, and if they do, then it is very brief and not reproducible [3].

Different research methods demand their own niche sample size calculations. Research methods that utilize natural language processing (NLP) are no exception. NLP is a branch of artificial intelligence within computer science that combines computational linguistics with statistical and machine learning models, enabling the analysis and processing of text data [6]. NLP methods are widely used within healthcare research due to the growing volume of data available from electronic health record (EHR) databases [7]. EHRs within a single hospital could generate about 150,000 pieces of data [7], some of which may be in textual form. Text is often used for ease of capturing fine-grained details or supplemental information which don't fit into any predetermined structured fields, such as patients' medical histories, preliminary diagnoses, medications, and so on [8].

A prerequisite for a good sample size is the availability of sufficient data to represent the larger population and provide adequate precision in output. This can be quite challenging in the healthcare domain due to the scarcity of openly available healthcare datasets, and the privacy regulations surrounding the use of such data from hospitals and primary care. Data scarcity is compounded by the fact that NLP-based supervised machine learning models require large numbers of human-annotated (in the healthcare domain, ideally clinician-annotated) data as a prerequisite, which can be a limitation due to time and cost constraints [9]. Insufficient sample sizes within NLP can lead to algorithms that do not perform adequately. We reviewed 11 papers describing past i2b2/n2c2 challenges (spanning from 2006 to 2018) [10–20] widely considered benchmarks in the field of clinical NLP. We found that while sample sizes were described for all training and test sets, a wide range of sample sizes were used (from 288 to 1243), and no justification was provided as to why any of the sample sizes were chosen. This has been summarised in Table 1. We do not say this as criticism specific to these highly cited and regarded papers, but to illustrate that justifications of sample size are rarely given, even in the best clinical NLP studies. This further highlights the need for recommendations on sample sizes specific to this field, and the need for guidelines on what sample sizes are needed, given limited data, to build machine learning models that perform well.

| Year | Challenge | Paper | Sample size mentioned | Justification provided? |
|---|---|---|---|---|
| 2006 | Deidentification & Smoking | Identifying Patient Smoking Status from Medical Discharge Records [10] | 928 | No |
| 2007 | Deidentification & Smoking | Evaluating the State-of-the-Art in Automatic De-identification [21] | 889 | No |
| 2008 | Obesity | Recognizing Obesity and Comorbidities in Sparse Data [12] | 1237 | No |
| 2009 | Medication | Extracting medication information from clinical text [13] | 1243 | No |
| 2010 | Relations | 2010 i2b2/VA challenge on concepts, assertions, and relations in clinical text [14] | 826 | No |
| 2011 | Coreference | Evaluating the state of the art in coreference resolution for electronic medical records [11] | 978 | No |
| 2012 | Temporal Relations | Evaluating temporal relations in clinical text: 2012 i2b2 Challenge [16] | 310 | No |
| 2014 | Deidentification & Heart Disease | Practical applications for natural language processing in clinical research: The 2014 i2b2/UTHealth shared tasks [17] | 600 | No |
| 2016 | RDoC for Psychiatry | A natural language processing challenge for clinical records: Research Domains Criteria (RDoC) for psychiatry [18] | 1000 | No |
| 2018 | ADE & Medication Extraction | Advancing the state of the art in automatic extraction of adverse drug events from narratives [19] | 505 | No |
| 2018 | Clinical Trial Cohort Selection | New approaches to cohort selection [20] | 288 | No |

Table 1. i2b2/n2c2 challenges - whether justification was provided for the sample sizes used

When referring to machine learning models in general and within NLP, sample size calculations can be used at different stages, such as for training, validation, and testing. Previous work has been conducted to determine sample sizes for validation [9,22] and limited research has been published on general sample sizes for NLP [23]. Sordo et al. (2005) examine the effect of sample size on the accuracy of classification with three classification methods (Naive Bayes, Decision Trees, and Support Vector Machines) using narrative reports from a hospital, classifying the smoking status of patients [23]. They conclude that there is indeed a correlation between the size of the training set and the classification rate, and models show improved performance when they are trained with bigger samples [23]. Using EHR databases, a recent study by Liu et al. (2021) highlighted the importance of selecting a sample that is unbiased and truly representative of the population in order to ensure high quality research [24]. While the work reported here does not address issues of bias due to the methodologies used to select samples, it does aim to extend previous research and explore the impact of sample sizes as well as class proportions for a binary classification task.

A substantial proportion of clinical decision making is dependent on risk-prediction models for health outcomes, which is why the margin of error for such models should be very low and the performance of such models should be thoroughly validated [25]. Pavlou et al. (2021) investigate the sample size requirements for such validation studies on prediction models [25]. A number of suggestions have been made on the appropriate sample sizes for such validation studies. These include at least 100 events in the validation data [26], or at least 100 events and 100 non-events [27]. However, such rules of thumbs are problematic and do not take into account the model or validation setting [22]. In response to this, Riley et al. (2020) suggest sample size calculations that incorporate other measures of model performance (such as expected c-statistics and calibration slope) which allows for more tailored sample size calculations based on the models of interest [22]. The aims for planning a sample size can vary, such as whether the sample size will be sufficient to reach a particular performance metric, or to detect differences in performance measures and pre-specified values, or both. This simulation focuses on the former, where the focus is on the performance metrics for each sample size and class proportion variation, while Pavlov et al. (2021) focused on the latter [25].

While methods such as power analysis, which are based on strong assumptions, are frequently used in statistical studies (such as prediction modelling) to determine appropriate sample sizes, with the general intention being the larger the sample size the more power associated with the study [28], this approach is not transferable to NLP approaches, and has therefore been underutilised within NLP [29]. This could be because of the nature of NLP data which does not conform to the standard experiment designs that are used in other studies [29–31], as also shown in complex statistical modelling [32]. Determining sample size in NLP applications is also complicated by the common use of the pre-trained models such as BERT [33], which convert text into a numerical representation, vectorising words in an embedding. These pre-trained models are based on large text corpora such as Wikipedia, or texts with specialised vocabularies, and are readily available in NLP software packages. Using embeddings in a local NLP project translates semantic knowledge of a large corpus to the local documents being classified [34], which can alter the required sample size or the choice of an optimal classifier. The recommendations made in this paper aim to complement such measures by providing some form of a standard that can be followed when building NLP models.

The performance indicators commonly used to evaluate and compare different NLP classification algorithms are AUC-ROC, precision, recall, accuracy, and F1-scores [35]. AUC-ROC is independent of specific thresholds or cut-offs [36], whereas metrics like precision, recall and accuracy are not. Therefore, F1-score and AUC-ROC will be the metrics to compare results from the different simulated classification models in this study.

The aim of this paper is to provide guidelines and suggestions on appropriate sample sizes for training NLP-based machine learned classification models. This was achieved by conducting straightforward simulations on an openly available healthcare dataset, utilising open-source software and widely used libraries, and building classifiers using varying sample sizes and class proportions as training data. Performances of the different models is compared and recommendations on sample sizes made based on this. The purpose is not to compare the different classifiers, but to recommend sample sizes based on their performances. Despite the simulations being straightforward, they yield valuable information. While some recommendations do exist within literature for other broader categories of research such as qualitative and quantitative research [3,4,37,38], and sample size for validation of machine learning models [9,22], predicting sample sizes using learning curve fitting [39–41], to the best our knowledge, such a simulation has not been reported for recommending sample sizes specific to NLP application development in the healthcare domain. This is of particular importance because NLP in the healthcare domain is often conducted on small datasets. The simulations may easily be extended for other parameters and use cases, to generate further recommendations. While these simulations have been conducted on a hypertension and diabetes diagnosis, the diagnosis is not what is being researched and has been used purely as an example. The objective is to understand how sample sizes and class proportions affect performance. We are not trying to generate any new knowledge on the diagnosis used.

# 2. Methods

A series of simulations were conducted on free-text healthcare data from the MIMIC-III database [42]. In the simulations, we varied different features of the NLP process, such as amount of training data, type of classifier algorithm, and prevalence of each class.

## 2.1 Data Source

Medical Information Mart for Intensive Care (MIMIC-III) is an EHR database which was developed by the Massachusetts Institute of Technology (MIT), and made available for researchers under a specified governance model [42]. MIMIC-III contains data on over 58,000 hospital admissions for over 45,000 patients, including about 1.2 million de-identified clinical notes, such as nursing and physician notes, discharge summaries, and ECG/radiology reports [42].

MIMIC-III was chosen for this study due to ease of access, thereby making the study reproducible. MIMIC-III is commonly used in healthcare research [43–46].

## 2.2 Ethics and Data Access

Access to the MIMIC-III database requires that the data be handled with care and respect as it contains detailed information about the clinical care of patients. Access was formally requested and granted through the processes documented on the MIMIC-III website[1]. A course protecting human research participants, including HIPAA (Health Insurance Portability and Accountability Act) requirements was completed. A data use agreement outlining appropriate data usage and standards of security was submitted.

[1] https://physionet.org/content/mimiciii/1.4/

## 2.3 Data Selection

For the simulations, we designed a simple binary classification task to classify documents within a subset of the database as coming from patients with a particular diagnosis or not. The most common diagnosis code within the database was identified as Unspecified Essential Hypertension, NOS (HTN, ICD-9 code 4019), and made up for 37% of patients within the MIMIC database. The reason for choosing a diagnosis that was so common in the database was so we would have enough data on the cases and non-cases for that diagnosis in order to run the simulation. However, to test the transferability of this approach, a less common diagnosis code was also used, that of diabetes mellitus without mention of complication (diabetes, ICD-9 code 25000), which made up for 15% of patients within the database. These codes were extracted from the “icd9_code” diagnosis column within the “d_icd9_diagnoses" table. None of the other tables within the database contain any diagnosis information. While the admissions table mentions diagnosis, this is not coded i.e., it is mentioned within the free-text and not considered to be the final diagnosis.

The simulated task would therefore be to classify documents as coming from patients diagnosed with HTN or not, and diabetes or not.

A SQL query was run to extract diagnosis, demographics and documents from patients who had the diagnosis of ICD-9 code 4019 (i.e., HTN), along with another subset of patients who did not have the diagnosis of 4019. The initial extraction consisted of 20,000 records from each subset. This is because there were about 20,000 patients with the diagnosis of 4019, and in order to match this, the same number of patients without this diagnosis code were extracted. A random sample of 5000 was selected for each class. The demographic and document distributions were compared for both classes to ensure they were similar to each other. We assume that this similarity, and the fact that they come from the same dataset, would ensure minimal noise in the data. This process was repeated for patients with and without diagnosis of ICD-9 code 25000 (i.e., diabetes) (Figure 2).

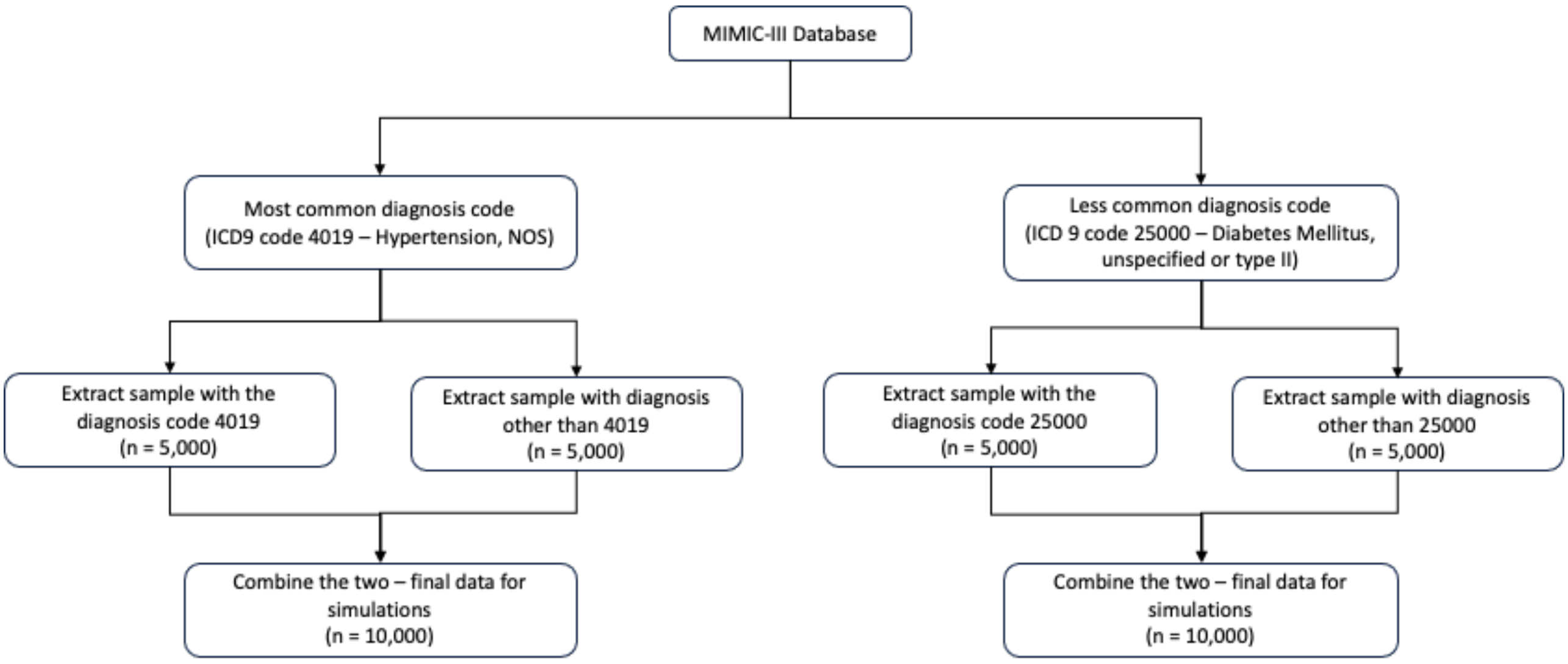

Figure 2. Extraction Plan

The majority of the notes within the MIMIC-III database are contained within the ‘note-events’ table (clinical notes such as nursing and physician notes, ECG reports, radiology reports, and discharge summaries) [42]. Clinical notes were extracted for both subsets of patients, and the distribution of document lengths calculated and compared.

A typical document within the combined dataset contains information such as dates (admission date, discharge date), history of present illness (explanation of how the patient's current illness developed and any treatment already undertaken), findings from physical examinations, reporting on various tests (scans, blood tests), any diagnosis and treatment plans.

## 2.4 Classifier

For the binary classification task, the diagnosis codes are considered to be the two classes, where patients with a diagnosis code of “4019” are categorised into class 1, and others without this diagnosis code as “class 0”. Similarly, for diabetes, patients with a diagnosis code of “25000” are categorised into class 1, and others without this diagnosis code as “class 0”. This approach for the simulation has been chosen in order to avoid the time-consuming task of manual annotation. The diagnosis codes for patients categorised as class 0 were examined to ensure this group did not contain any diagnosis similar to hypertension or diabetes (a list of all diagnosis codes for class 0 is given in appendix B). Thirteen documents within the class 0 of the HTN subset contained hypertension-related diagnosis, such as surgical complication-hypertension (3 patients), malignant hypertension (7 patients), primary pulmonary hypertension (2 patients), and portal hypertension (1 patient). These were removed from the dataset to give the final dataset that was used to run the simulations with varying features as outlined in table 2.

Before building the various classifier models, some pre-processing of the text data was undertaken using the Python NLTK (Natural Language Toolkit) library [47]. All text was converted to lowercase, any trailing whitespaces, markups such as HTML tags and punctuations/symbols were removed. Common stop words were removed. Words were lemmatised and tokenised. The Python library sklearn [48] was used for the non-BERT classifiers, and the Python Huggingface transformers library (version 4.5.0) [49] for the BERT model. The pre-trained BERT_base model was fine-tuned on the simulation data. No steps have been undertaken to balance the data when the class proportions are imbalanced (such as 99/1, 95/5, 90/10, 80/20 class proportions) as the aim was to investigate the effect of this imbalance.

Each sample was split into train/test/validation sets in the proportions of 60/20/20. The class proportions for each set followed the same prevalence as the main sample size. For example, a sample size of 600 with a 50/50 split would contain 300 documents each in both classes. This was replicated within the train, test and validation sets so the classes were evenly distributed. Distribution of words within each class in the sets were also compared to the overall sample to ensure a similar distribution was represented throughout. For the BERT model, training and validation loss were measured to ensure no overfitting of the models.

| Variables | Examples |
|---|---|
| Size of sample | 5000, 4000, 3000, 2000, 1000, 800, 600, 500, 400, 200 |
| Type of classifier/algorithm | Logistic Regression (LR) [50],<br>Naive Bayes (NB) [51],<br>Random Forest (RF) [52],<br>Decision Tree (DT) [52],<br>Support Vector Classifier (SVC) [53],<br>Linear Support Vector Classifier (LSVC) [53],<br>Stochastic Gradient Descent (SGD) [54],<br>K-Nearest Neighbour (KNN) [55],<br>BERT (BERT_base) [33] |
| Prevalence of each label | 99/1, 95/5, 90/10, 80/20, 70/30, 60/40, 50/50 |

Table 2. Features to be varied

The weighted average of F1-score (and confidence intervals) and AUC score were calculated upon varying different features within this classification task. Sample size recommendations have been made, based on the combination of features that perform best.

The queries and code have been made available on GitHub[2].

# 3. Results

## 3.1 Data Summary

MIMIC-III contains 46,520 patients and 58,976 admission records. HTN (ICD-9 code 4019) makes up for 37% (17,613) of all patients and 53% (24,719) of all admissions within the database. Diabetes (ICD-9 code 25000) makes up for 15% (7,370) of all patients and 24% (11,183) of all admissions within the database.

The demographic and document distributions were compared to ensure they were similar enough to each across both classes in the respective subset groups i.e., HTN and diabetes. The demographic distributions are outlined in Table 4.

| **Demographic** | **HTN cohort** | | **Diabetes cohort** | |
|---|---|---|---|---|
| | **Class 0** | **Class 1** | **Class 0** | **Class 1** |
| Gender | Male: 55%<br>Female: 45% | Male: 59%<br>Female: 41% | Male: 56%<br>Female: 44% | Male: 62%<br>Female: 38% |
| Age* | Mean: -61<br>Min: -169<br>Max: 221 | Mean: -55<br>Min: -151<br>Max: 221 | Mean: -68<br>Min: -179<br>Max: 221 | Mean: -58<br>Min: -138<br>Max: 220 |
| Ethnicity | White: 66%<br>Unknown: 12%<br>Black: 9%<br>Hispanic/Latino: 4%<br>Multi-race/Other: 4% | White: 68%<br>Unknown: 14%<br>Black: 9%<br>Hispanic/Latino: 4%<br>Multi-race/Other: 3% | White: 67%<br>Unknown: 10%<br>Black: 10%<br>Hispanic/Latino: 6%<br>Multi-race/Other: 3% | White: 60%<br>Unknown: 17%<br>Black: 9%<br>Hispanic/Latino: 7%<br>Multi-race/Other: 3% |

[2] https://github.com/jayachaturvedi/sample_size_in_healthcare_NLP

| | | | | |
|---|---|---|---|---|
| | Asian: 5% | Asian: 2% | Asian: 4% | Asian: 4% |

Table 4. Demographic distributions for both classes in both cohorts
*Ages within MIMIC-III may be negative. For the purposes of de-identification and keeping in line with the HIPAA regulations, the database providers have shifted any dates within the database (including age) into the future by a random offset for each individual patient. This has been done in a consistent manner so that the intervals between stays and discharge are preserved. However, due to this shift, hospital stays appear to occur between the years 2100 and 2200, leading to calculated ages being negative. For patients with a date of birth over 89, their age within the database appears as being over 100 years old [42].

Some patients within class 1 contained hundreds of documents, and so have been eliminated by the application of a limit to the number of documents per patient, in order to maintain consistency. This limit was set to less than or equal to 50 documents per patient. After applying the threshold of 50 documents, the document distribution is displayed in Figures 3a for HTN and 3b for diabetes cohort.

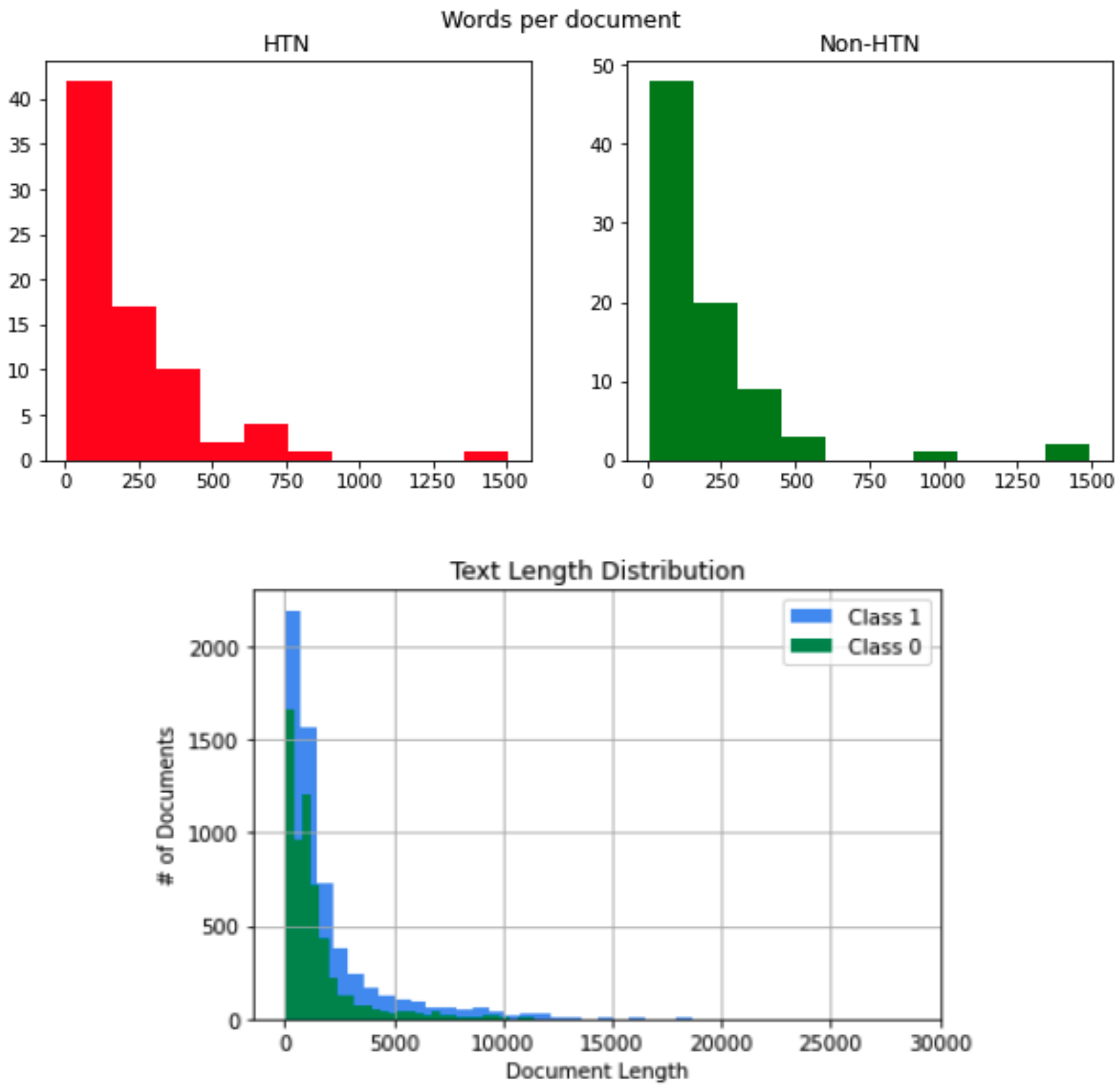

Figure 3a. Document distribution between the two classes - Hypertension

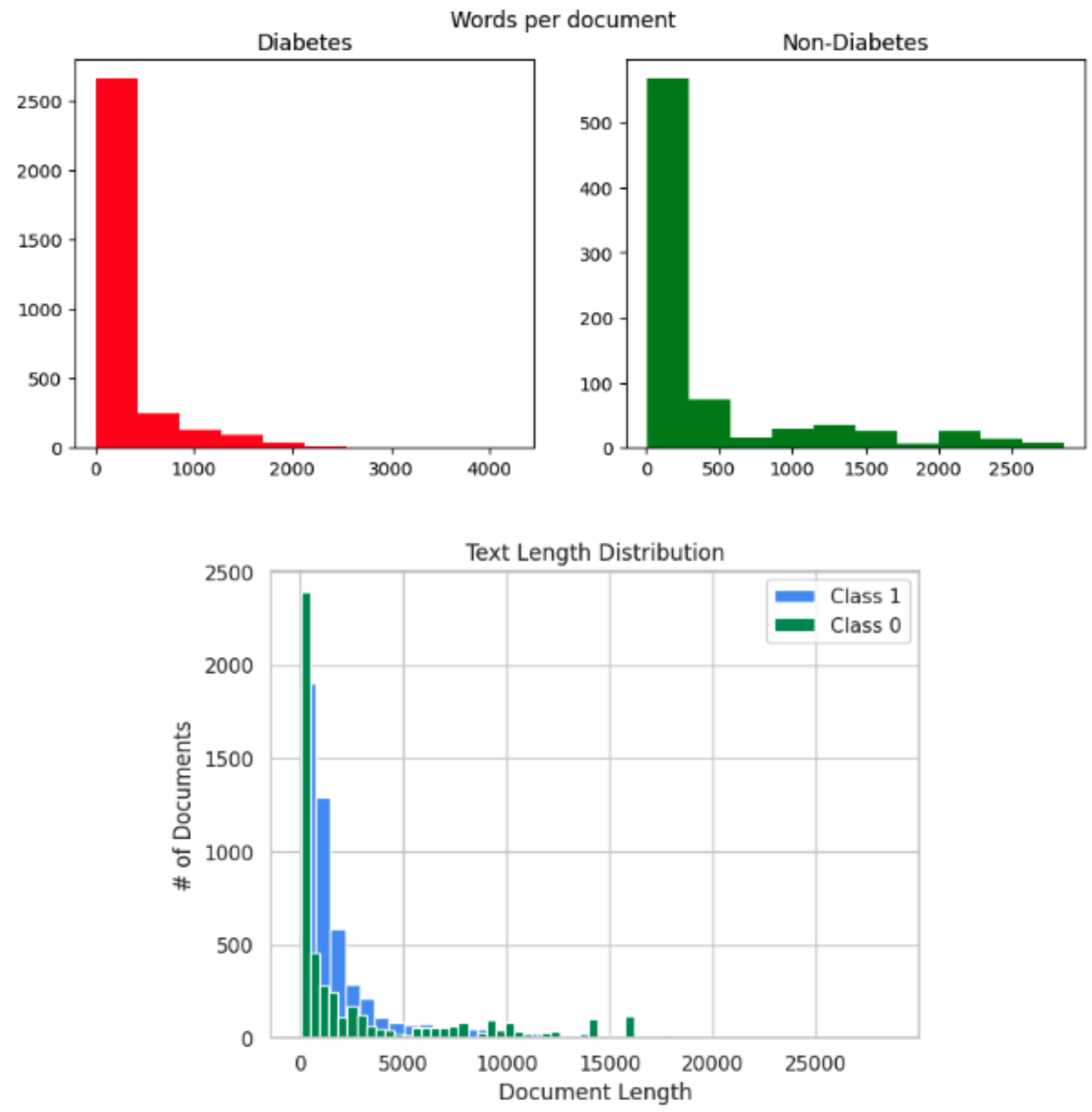

Figure 3b. Document distribution between the two classes - Diabetes

## 3.2 Classifier

### Effect of variations on Classifier Performance

Figure 4 shows the impact of variations in sample size and classifiers on one class proportion variation (50/50).

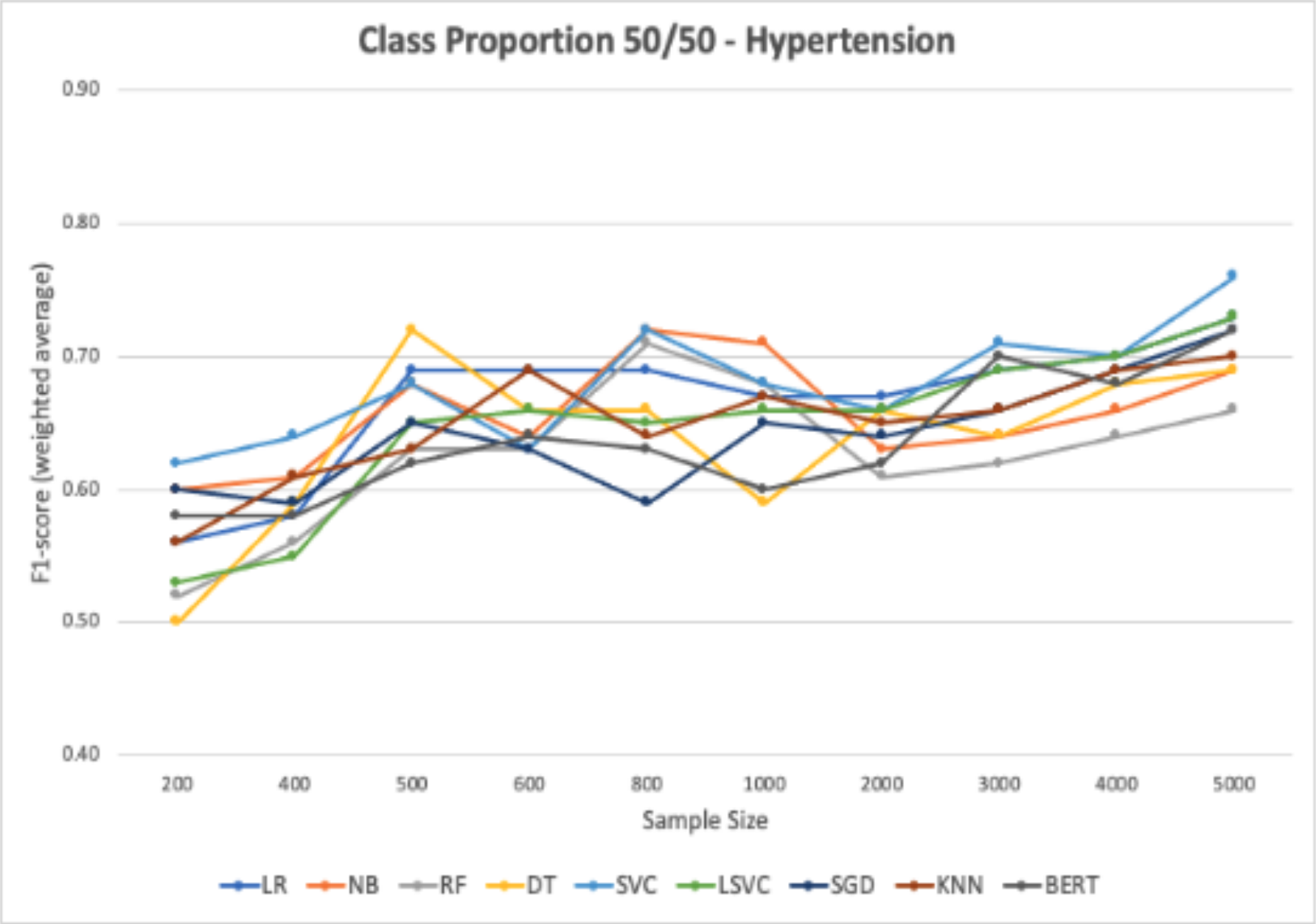

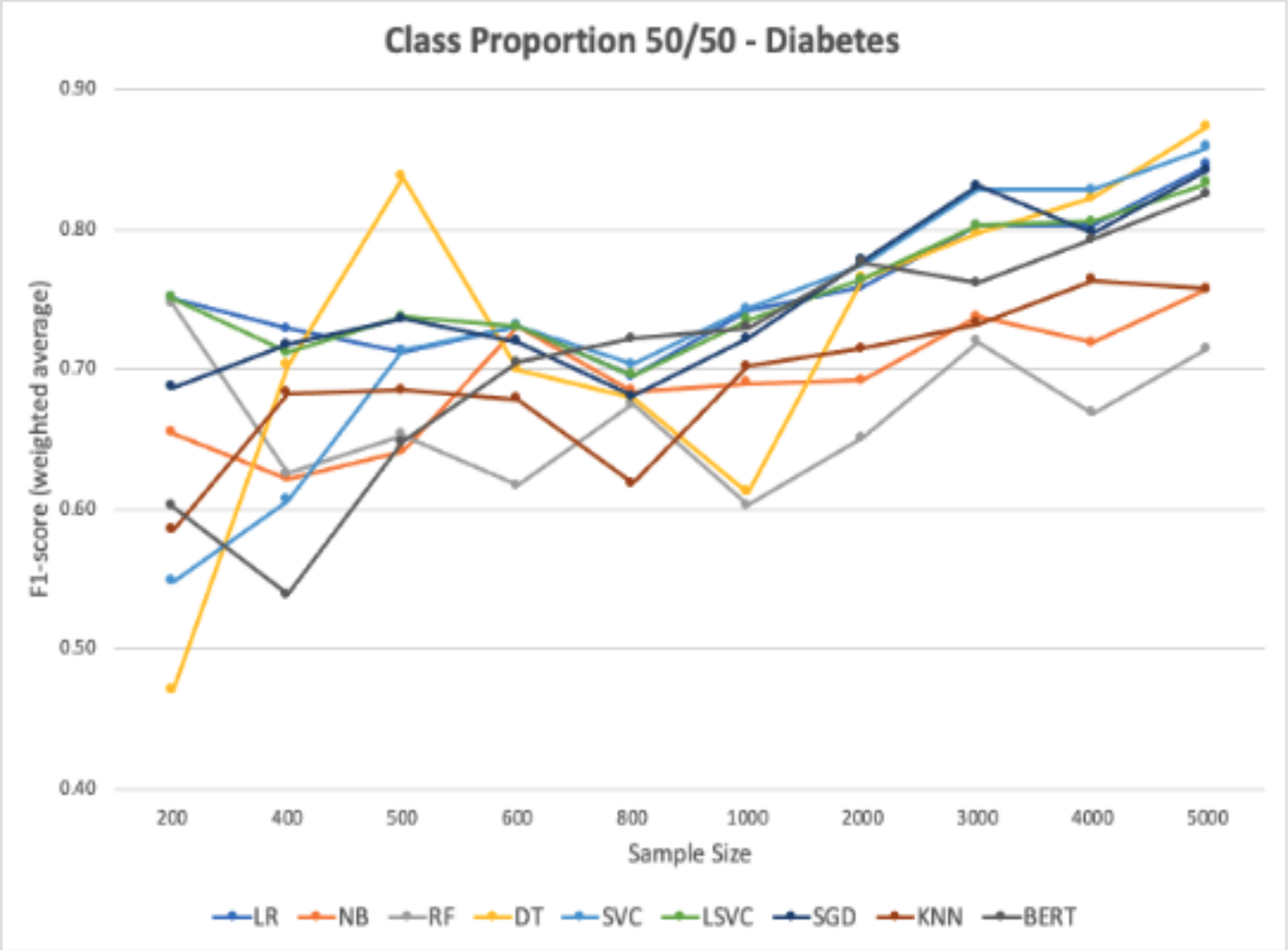

Figure 4. F1 for each classifier and different sample sizes at 50/50 class proportion - HTN and diabetes

A table showing the metrics for all the classifiers and sample size/class proportion variations for both the diagnosis subsets is given in Appendix A and C, with the range of AUC and F1 scores for some sample sizes and class proportions of the HTN and Diabetes subsets summarised in Table 5. The best performing classifier given in this table was based on the F1 score. AUC scores lead to a different classifier performance ranking, and can be found in Appendix A.

| **Range of AUC and F1 scores<br>(Range of AUC [classifier with highest AUC]** |
|---|

| Range of F1 scores [classifier with highest F1 score]) | | | | | |
|---|---|---|---|---|---|
| **Hypertension** | | | | | |
| **Sample Size** | **Class Proportion** | | | | |
| | **90/10** | **80/20** | **70/30** | **60/40** | **50/50** |
| **5000** | 0.63 - 0.73 [LR]<br>0.87 - 0.91 [LSVC] | 0.65 - 0.78 [LR]<br>0.76 - 0.83 [SVC] | 0.67 - 0.76 [SVC]<br>0.67 - 0.78 [LR] | 0.67 - 0.82 [SVC]<br>0.63 - 0.76 [SVC] | 0.69 - 0.83 [SVC]<br>0.66 - 0.76 [SVC] |
| **3000** | 0.61 - 0.74 [SVC]<br>0.85 - 0.91 [LSVC] | 0.60 - 0.75 [LR]<br>0.74 - 0.81 [LR] | 0.66 - 0.77 [LSVC]<br>0.66 - 0.78 [LSVC] | 0.62 - 0.75 [LSVC]<br>0.59 - 0.71 [BERT] | 0.63 - 0.77 [SVC]<br>0.61 - 0.71 [SVC] |
| **1000** | 0.50 - 0.75 [SVC]<br>0.79 - 0.88 [BERT] | 0.54 - 0.72 [SGD]<br>0.74 - 0.83 [SGD] | 0.53 - 0.63 [RF]<br>0.64 - 0.72 [SVC] | 0.60 - 0.71 [SGD]<br>0.58 - 0.67 [SGD] | 0.59 - 0.74 [LR]<br>0.59 - 0.71 [NB] |
| **500** | 0.43 - 0.61 [KNN]<br>0.85 - 0.94 [KNN] | 0.59 - 0.78 [LR]<br>0.67 - 0.86 [LSVC] | 0.59 - 0.71 [NB]<br>0.72 - 0.76 [LR] | 0.64 - 0.73 [SVC]<br>0.58 - 0.71 [LR] | 0.59 - 0.77 [LR]<br>0.62 - 0.72 [DT] |
| **200** | 0.30 - 0.63 [NB]<br>0.85 - 0.91 [KNN] | 0.48 - 0.65 [DT]<br>0.58 - 0.84 [KNN] | 0.31 - 0.51 [NB]<br>0.55 - 0.73 [SVC] | 0.47 - 0.78 [SVC]<br>0.40 - 0.74 [BERT] | 0.34 - 0.68 [SGD]<br>0.35 - 0.62 [NB] |
| **Diabetes** | | | | | |
| **5000** | 0.68 - 0.87 [LR]<br>0.86 - 0.93 [LSVC] | 0.72 - 0.88 [LR]<br>0.75 - 0.88 [SGD] | 0.79 - 0.91 [SVC]<br>0.64 - 0.87 [SVC] | 0.79 - 0.92 [SVC]<br>0.64 - 0.86 [SVC] | 0.81 - 0.93 [SVC]<br>0.71 - 0.87 [DT] |
| **3000** | 0.58 - 0.88 [LR]<br>0.87 - 0.93 [LSVC] | 0.70 - 0.85 [LR]<br>0.75 - 0.87 [SGD] | 0.74 - 0.88 [LR]<br>0.67 - 0.84 [SVC] | 0.76 - 0.90 [SVC]<br>0.66 - 0.84 [SVC] | 0.76 - 0.90 [SVC]<br>0.72 - 0.83 [SGD] |
| **1000** | 0.50 - 0.84 [LR]<br><br>0.83 - 0.89 [SGD] | 0.50 - 0.80 [LR]<br><br>0.68 - 0.76 [KNN] | 0.73 - 0.87 [RF]<br><br>0.71 - 0.81 [KNN] | 0.70 - 0.84 [LR]<br><br>0.59 - 0.78 [SGD] | 0.61 - 0.85 [LSVC]<br><br>0.60 - 0.74 [LSVC] |
| **500** | 0.50 - 0.79 [RF]<br><br>0.72 - 0.88 [SGD] | 0.59 - 0.83 [RF]<br><br>0.75 - 0.87 [KNN] | 0.52 - 0.86 [LSVC]<br><br>0.50 - 0.78 [SGD] | 0.62 - 0.86 [LR]<br><br>0.62 - 0.80 [SGD] | 0.67 - 0.84 [DT]<br><br>0.64 - 0.84 [DT] |
| **200** | 0.45 - 0.79 [NB]<br><br>0.80 - 0.85 [BERT] | 0.50 - 0.83 [LR]<br><br>0.63 - 0.86 [KNN] | 0.50 - 0.89 [RF]<br><br>0.45 - 0.84 [LSVC] | 0.59 - 0.84 [LR]<br><br>0.36 - 0.69 [KNN] | 0.47 - 0.87 [RF]<br><br>0.47 - 0.75 [LSVC] |

Table 5. Range of AUC and F1 scores for each sample size and class proportion

# 4. Discussion

The simulations conducted in this project are aimed at providing some recommendations on sample sizes that will be useful when building a classifier, based on class proportions and classifier types. As expected, classifiers built with larger sample sizes showed better performance in general. The results showed that larger sample sizes resulted in some classifiers performing better, and others with more balanced classes. For example, smaller samples (800 and below) resulted in better performance by the KNN classifier and more frequently with imbalanced class proportions such as 90/10, 80/20 and 70/30, when compared to larger samples (1000 and above) that generated better performance with the BERT model. This might be because KNN is a distance-based algorithm and is able to calculate distances within small datasets with ease [56]. KNNs are known to not perform well with large datasets due to the curse of dimensionality where the distance functions that are used within KNNs are rendered ineffective due to high dimensionality within larger datasets [57]. Apart from this, the computational costs of calculating distances between new points within the dataset is very high too [56]. For some classifiers, such as SVC and LSVC, larger samples are required in order to obtain good performance results. This is interesting because SVCs are generally not considered ideal for large datasets because they are slow to train when using large datasets that also contain large number of features and variations, making it computationally infeasible [58]. Imbalanced classes performed best most frequently with the BERT model. This could be because the model utilises transfer learning and being pre-trained on a large corpus of language, its architecture produces pre-trained context-dependent embeddings which encode aspects of general language, as shown by the fact that they have proven powerful in solving a multitude of NLP tasks, including handling imbalanced classification tasks without need for any further augmentation or manipulation [59]. Finally, although we compared a number of algorithms, our list is far from being exhaustive, and future work may include a wider range of deep learning and penalised regression models.

As the sample sizes reduce within the simulations, the confidence intervals get wider (reported in Appendix A and C), as expected. Similarly, the AUC scores are lower with the smaller sample sizes compared to larger ones and often approach on average chance classification (AUC=0.5) i.e., it cannot separate between classes [60]. This is attributed to the accuracy of the classifier rather than the data (sample size and class proportions). Since the accuracy of a classifier is affected by sample size and class proportions, it in turn affects the AUC score.

An important consideration in creation of labelled training data in the real world is the human annotation process, and inter-annotator agreements (IAA). This work does not account for IAA scores, since we are using the diagnosis code as the label, which was potentially assigned by a single coder within the hospital and is being used as the gold standard in our work. An assessment of clinical coding within routinely collected hospital data was conducted by Dixon et al. (1998) which found that inter-coder agreement varied between different medical conditions [61]. While this work has focussed on sample sizes, and not addressed the issues of IAA, they are both important factors that might determine the performance of an NLP classifier.

Lastly, the data heterogeneity within healthcare text is a known challenge when it comes to transferability of results and determining whether the recommendations made in this paper can be applied to data from another healthcare database. Since the results obtained in this paper were from data of a critical care unit based in the United States, there will be differences in the structure and content of the textual data when compared to other sources of health data. Even within the same database, the results might vary when a classification task is conducted on another diagnosis code. However, the vocabulary used within different healthcare datasets would have some overlap due to the common terminologies used, so these results can prove to be a useful guide.

# 5. Conclusion

This paper provides recommendations on sample size for training data when building classification models. These recommendations are based on simulations that were conducted on the MIMIC-III dataset, using patient documents with the most common diagnosis code (HTN) as class 1 and a similar cohort of patient documents with any other diagnosis code as class 0. The sample sizes were varied incrementally from 200 to 5000 documents, and class proportions varied from a 50/50 split of classes to a 90/10 split. Different classification algorithms were used on these varying sample sizes and class proportions. This simulation was repeated with a less common diagnosis code (diabetes), as a test of the transferability of this approach. The results have been reported briefly in Table 4 and in more detail in Appendix A. The objective is to use these recommendations as guidelines when conducting similar classification tasks within the healthcare domain.

While it is not unusual for reports of clinical NLP to compare results for different parameters and algorithms with respect to specific narrow research questions, to the best of our knowledge this paper is the first to report a reproducible methodology and guidelines using open tools and data, for the purpose of guiding general decisions on sample size across a range of NLP research. The recommendations from this work could be applicable to classification tasks being conducted on other similar datasets within healthcare. One of the limitations of this work is that this was carried out on a critical care unit database which contains specific terminologies and potentially more abbreviations within their notes, which might not be as transferable to a different kind of dataset, such as health records from other medical specialties, which might contain different styles and complexities of narrative. However, the methodology presented here can be replicated by other researchers for sample size estimations using their own datasets. An understanding of appropriate sample size will enable researchers to better judge both replicability and reproducibility of reported studies, and therefore to understand the limitations of those studies.

In future work, we intend to vary the level of classification (such as sentence or token level vs. document level). This would have to overcome the lack of availability of labelled health record data for some levels, such as tokens. We will also consider the split between train/test/validation sets within a sample, and the type of classification (such as multiclass vs. binary). We plan to run a simulation on a different dataset to test the transferability of our approach. We also plan to expand our simulation approach to investigate if such text features as size of the underlying vocabulary, number of words per document, or similarity of the descriptive words for the positive and negative classes, affect the minimum training sample size required for an NLP model. The methodology proposed here has provided guidelines and recommendations on what sample sizes and class proportions should be used for binary health record document classification. The same methodology could be used for future extensions to this work.

# 6. Authors' Contributions

The idea was conceived by JC, AR, DS, RS and DSham. JC conducted the simulations and drafted the manuscript. AR, RS, DS, DSham, SV, and FZ provided guidance in the design and interpretation of results. All authors commented on drafts of the manuscript and approved the final version.

# 7. Funding Statement

AR was part-funded by Health Data Research UK, an initiative funded by UK Research and Innovation, Department of Health and Social Care (England) and the devolved administrations, and leading medical research charities. AR's and RS's salaries were part supported by the UK Prevention Research Partnership (Violence, Health and Society; MR-VO49879/1), an initiative funded by UK Research and Innovation Councils, the Department of Health and Social Care (England) and the UK devolved administrations, and leading health research charities. DS, RS and AR are part-funded by the National Institute for Health Research (NIHR) Biomedical Research Centre at South London and Maudsley NHS Foundation Trust and King's College London. RS is additionally part-funded by the National Institute for Health Research (NIHR) Applied Research Collaboration South London (NIHR ARC South London) at King's College Hospital NHS Foundation Trust, and by the DATAMIND HDR UK Mental Health Data Hub (MRC grant MR/W014386). FZ was funded by the Swiss National Science Foundation (SNSF grant 188929). JC was supported by the KCL funded Centre for Doctoral Training (CDT) in Data-Driven Health. DSh is funded by the King's College London Biostatistics and Health Informatics PhD studentship. The funders had no role in study design, data collection and analysis, decision to publish, or preparation of the manuscript.

## Declaration of Competing Interests

RS declares research support received in the last 3 years from Janssen, GSK and Takeda. AR declares research support received in the last 3 years from Takeda and RSM UK. The other authors have no competing interests to declare.

**Appendix A - HTN**
All classifiers, sample sizes, class proportions and scores

| Classifier | Sampl e Size | Class Proportion (class 1/ class 0) | F1-score Weighted avg. (95% CI) | AUC Score |
|---|---|---|---|---|
| Logistic Regression | 200 | 90/10 | 0.91 (1-0.77) | 0.33 |
| | | 80/20 | 0.79 (0.95-0.64) | 0.5 |
| | | 70/30 | 0.65 (0.84-0.47) | 0.38 |
| | | 60/40 | 0.58 (0.77-0.39) | 0.69 |
| | | 50/50 | 0.53 (0.71-0.35) | 0.66 |
| | 400 | 90/10 | 0.89 (0.98-0.79) | 0.67 |
| | | 80/20 | 0.73 (0.87-0.6) | 0.65 |
| | | 70/30 | 0.74 (0.87-0.6) | 0.62 |
| | | 60/40 | 0.61 (0.76-0.46) | 0.56 |
| | | 50/50 | 0.58 (0.7-0.46) | 0.68 |
| | 500 | 90/10 | 0.93 (0.96-0.9) | 0.5 |
| | | 80/20 | 0.85 (0.93-0.76) | 0.78 |
| | | 70/30 | 0.76 (0.86-0.64) | 0.66 |
| | | 60/40 | 0.71 (0.81-0.58) | 0.68 |
| | | 50/50 | 0.69 (0.79-0.59) | 0.77 |
| | 600 | 90/10 | 0.8 (0.89-0.7) | 0.57 |
| | | 80/20 | 0.81 (0.9-0.72) | 0.71 |
| | | 70/30 | 0.76 (0.86-0.65) | 0.71 |
| | | 60/40 | 0.64 (0.75-0.55) | 0.68 |
| | | 50/50 | 0.69 (0.78-0.59) | 0.8 |
| | 800 | 90/10 | 0.92 (0.97-0.85) | 0.65 |
| | | 80/20 | 0.84 (0.91-0.76) | 0.65 |
| | | 70/30 | 0.77 (0.85-0.69) | 0.71 |
| | | 60/40 | 0.65 (0.73-0.57) | 0.75 |
| | | 50/50 | 0.69 (0.77-0.61) | 0.72 |
| | 1000 | 90/10 | 0.82 (0.89-0.75) | 0.72 |
| | | 80/20 | 0.82 (0.89-0.74) | 0.71 |
| | | 70/30 | 0.7 (0.77-0.62) | 0.62 |
| | | 60/40 | 0.66 (0.74-0.58) | 0.68 |
| | | 50/50 | 0.67 (0.74-0.6) | 0.74 |
| | 2000 | 90/10 | 0.88 (0.92-0.83) | 0.66 |
| | | 80/20 | 0.82 (0.87-0.78) | 0.74 |
| | | 70/30 | 0.76 (0.81-0.71) | 0.75 |
| | | 60/40 | 0.72 (0.78-0.67) | 0.78 |
| | | 50/50 | 0.67 (0.72-0.62) | 0.74 |
| | 3000 | 90/10 | 0.91 (0.93-0.87) | 0.73 |
| | | 80/20 | 0.81 (0.85-0.77) | 0.75 |
| | | 70/30 | 0.77 (0.81-0.73) | 0.77 |
| | | 60/40 | 0.68 (0.73-0.63) | 0.75 |
| | | 50/50 | 0.69 (0.74-0.65) | 0.76 |
| | 4000 | 90/10 | 0.89 (0.92-0.86) | 0.7 |
| | | 80/20 | 0.81 (0.84-0.77) | 0.76 |
| | | 70/30 | 0.77 (0.8-0.74) | 0.79 |
| | | 60/40 | 0.71 (0.75-0.68) | 0.77 |
| | | 50/50 | 0.7 (0.73-0.66) | 0.78 |
| | 5000 | 90/10 | 0.91 (0.93-0.89) | 0.73 |
| | | 80/20 | 0.83 (0.86-0.8) | 0.78 |
| | | 70/30 | 0.78 (0.81-0.74) | 0.74 |
| | | 60/40 | 0.75 (0.78-0.72) | 0.81 |
| | | 50/50 | 0.73 (0.76-0.7) | 0.82 |
| Decision Tree | 200 | 90/10 | 0.88 (0.98-0.74) | 0.47 |
| | | 80/20 | 0.82 (0.94-0.67) | 0.65 |
| | | 70/30 | 0.55 (0.74-0.39) | 0.37 |
| | | 60/40 | 0.45 (0.65-0.28) | 0.47 |
| | | 50/50 | 0.53 (0.69-0.35) | 0.54 |
| | 400 | 90/10 | 0.89 (0.97-0.8) | 0.58 |
| | | 80/20 | 0.73 (0.84-0.62) | 0.61 |
| | | 70/30 | 0.74 (0.85-0.61) | 0.64 |
| | | 60/40 | 0.5 (0.63-0.38) | 0.47 |
| | | 50/50 | 0.49 (0.6-0.37) | 0.49 |
| | 500 | 90/10 | 0.88 (0.92-0.83) | 0.51 |
| | | 80/20 | 0.72 (0.81-0.62) | 0.59 |
| | | 70/30 | 0.72 (0.81-0.6) | 0.63 |
| | | 60/40 | 0.66 (0.77-0.56) | 0.64 |
| | | 50/50 | 0.72 (0.81-0.63) | 0.73 |
| | 600 | 90/10 | 0.78 (0.87-0.68) | 0.49 |
| | | 80/20 | 0.75 (0.84-0.65) | 0.66 |
| | | 70/30 | 0.77 (0.86-0.67) | 0.71 |
| | | 60/40 | 0.55 (0.64-0.45) | 0.54 |
| | | 50/50 | 0.66 (0.76-0.56) | 0.67 |
| | 800 | 90/10 | 0.9 (0.95-0.84) | 0.68 |
| | | 80/20 | 0.8 (0.88-0.72) | 0.62 |
| | | 70/30 | 0.76 (0.83-0.68) | 0.7 |
| | | 60/40 | 0.63 (0.71-0.54) | 0.62 |
| | | 50/50 | 0.66 (0.73-0.58) | 0.66 |
| | 1000 | 90/10 | 0.79 (0.86-0.72) | 0.5 |
| | | 80/20 | 0.76 (0.82-0.68) | 0.62 |
| | | 70/30 | 0.64 (0.72-0.56) | 0.55 |
| | | 60/40 | 0.66 (0.74-0.59) | 0.64 |
| | | 50/50 | 0.59 (0.68-0.51) | 0.59 |
| | 2000 | 90/10 | 0.84 (0.88-0.8) | 0.63 |
| | | 80/20 | 0.77 (0.81-0.72) | 0.67 |
| | | 70/30 | 0.74 (0.79-0.68) | 0.66 |
| | | 60/40 | 0.64 (0.69-0.59) | 0.63 |
| | | 50/50 | 0.66 (0.71-0.61) | 0.67 |
| | 3000 | 90/10 | 0.85 (0.88-0.82) | 0.61 |
| | | 80/20 | 0.74 (0.78-0.69) | 0.6 |
| | | 70/30 | 0.71 (0.75-0.67) | 0.66 |
| | | 60/40 | 0.64 (0.68-0.59) | 0.62 |
| | | 50/50 | 0.64 (0.67-0.59) | 0.63 |
| | 4000 | 90/10 | 0.85 (0.87-0.82) | 0.63 |
| | | 80/20 | 0.76 (0.79-0.72) | 0.64 |
| | | 70/30 | 0.7 (0.74-0.67) | 0.68 |
| | | 60/40 | 0.67 (0.7-0.63) | 0.67 |
| | | 50/50 | 0.68 (0.71-0.65) | 0.68 |
| | 5000 | 90/10 | 0.87 (0.9-0.85) | 0.63 |
| | | 80/20 | 0.78 (0.81-0.75) | 0.65 |
| | | 70/30 | 0.73 (0.76-0.69) | 0.67 |
| | | 60/40 | 0.72 (0.75-0.69) | 0.71 |
| | | 50/50 | 0.69 (0.72-0.66) | 0.69 |
| K-Nearest Neighbour | 200 | 90/10 | 0.91 (1-0.77) | 0.38 |
| | | 80/20 | 0.84 (0.97-0.66) | 0.54 |
| | | 70/30 | 0.67 (0.84-0.51) | 0.37 |
| | | 60/40 | 0.54 (0.74-0.37) | 0.62 |
| | | 50/50 | 0.57 (0.72-0.39) | 0.59 |
| | 400 | 90/10 | 0.9 (0.98-0.81) | 0.53 |
| | | 80/20 | 0.73 (0.86-0.61) | 0.71 |
| | | 70/30 | 0.68 (0.81-0.54) | 0.45 |
| | | 60/40 | 0.56 (0.67-0.44) | 0.6 |
| | | 50/50 | 0.65 (0.76-0.53) | 0.65 |
| | 500 | 90/10 | 0.94 (0.96-0.91) | 0.61 |
| | | 80/20 | 0.85 (0.93-0.76) | 0.76 |
| | | 70/30 | 0.73 (0.84-0.62) | 0.59 |
| | | 60/40 | 0.66 (0.77-0.54) | 0.64 |
| | | 50/50 | 0.62 (0.73-0.51) | 0.74 |
| | 600 | 90/10 | 0.82 (0.91-0.72) | 0.63 |
| | | 80/20 | 0.82 (0.9-0.73) | 0.67 |
| | | 70/30 | 0.79 (0.87-0.69) | 0.73 |
| | | 60/40 | 0.6 (0.7-0.5) | 0.65 |
| | | 50/50 | 0.69 (0.78-0.59) | 0.74 |
| | 800 | 90/10 | 0.88 (0.93-0.81) | 0.72 |
| | | 80/20 | 0.82 (0.9-0.74) | 0.61 |
| | | 70/30 | 0.79 (0.87-0.71) | 0.64 |
| | | 60/40 | 0.64 (0.72-0.57) | 0.74 |
| | | 50/50 | 0.64 (0.73-0.55) | 0.72 |
| | 1000 | 90/10 | 0.86 (0.92-0.79) | 0.6 |
| | | 80/20 | 0.81 (0.89-0.74) | 0.64 |
| | | 70/30 | 0.67 (0.75-0.58) | 0.6 |
| | | 60/40 | 0.59 (0.67-0.51) | 0.61 |
| | | 50/50 | 0.67 (0.74-0.6) | 0.72 |
| | 2000 | 90/10 | 0.88 (0.92-0.83) | 0.61 |
| | | 80/20 | 0.82 (0.87-0.77) | 0.65 |
| | | 70/30 | 0.74 (0.79-0.68) | 0.71 |
| | | 60/40 | 0.72 (0.76-0.67) | 0.78 |
| | | 50/50 | 0.65 (0.7-0.6) | 0.73 |
| | 3000 | 90/10 | 0.9 (0.93-0.87) | 0.71 |
| | | 80/20 | 0.77 (0.82-0.73) | 0.69 |
| | | 70/30 | 0.74 (0.78-0.7) | 0.71 |
| | | 60/40 | 0.68 (0.72-0.64) | 0.71 |
| | | 50/50 | 0.66 (0.7-0.61) | 0.73 |
| | 4000 | 90/10 | 0.85 (0.88-0.82) | 0.65 |
| | | 80/20 | 0.81 (0.85-0.78) | 0.71 |
| | | 70/30 | 0.74 (0.77-0.7) | 0.76 |
| | | 60/40 | 0.66 (0.7-0.62) | 0.73 |
| | | 50/50 | 0.69 (0.73-0.66) | 0.77 |
| | 5000 | 90/10 | 0.9 (0.93-0.87) | 0.7 |
| | | 80/20 | 0.81 (0.84-0.78) | 0.72 |
| | | 70/30 | 0.76 (0.79-0.72) | 0.73 |
| | | 60/40 | 0.71 (0.74-0.68) | 0.77 |
| | | 50/50 | 0.7 (0.73-0.67) | 0.79 |
| Linear Support Vector Classifier | 200 | 90/10 | 0.91 (1-0.77) | 0.3 |
| | | 80/20 | 0.79 (0.94-0.6) | 0.48 |
| | | 70/30 | 0.64 (0.82-0.44) | 0.35 |
| | | 60/40 | 0.58 (0.78-0.4) | 0.7 |
| | | 50/50 | 0.49 (0.65-0.32) | 0.66 |
| | 400 | 90/10 | 0.88 (0.98-0.77) | 0.72 |
| | | 80/20 | 0.73 (0.86-0.57) | 0.64 |
| | | 70/30 | 0.73 (0.85-0.59) | 0.62 |
| | | 60/40 | 0.63 (0.77-0.5) | 0.57 |
| | | 50/50 | 0.58 (0.71-0.47) | 0.68 |
| | 500 | 90/10 | 0.93 (0.97-0.89) | 0.52 |
| | | 80/20 | 0.86 (0.94-0.77) | 0.76 |
| | | 70/30 | 0.76 (0.87-0.65) | 0.66 |
| | | 60/40 | 0.7 (0.8-0.58) | 0.69 |
| | | 50/50 | 0.65 (0.76-0.54) | 0.76 |
| | 600 | 90/10 | 0.8 (0.89-0.69) | 0.57 |
| | | 80/20 | 0.82 (0.9-0.72) | 0.69 |
| | | 70/30 | 0.76 (0.86-0.64) | 0.72 |
| | | 60/40 | 0.63 (0.72-0.52) | 0.68 |
| | | 50/50 | 0.67 (0.76-0.57) | 0.8 |
| | 800 | 90/10 | 0.92 (0.97-0.86) | 0.65 |
| | | 80/20 | 0.84 (0.9-0.75) | 0.64 |
| | | 70/30 | 0.78 (0.86-0.69) | 0.72 |
| | | 60/40 | 0.66 (0.75-0.57) | 0.75 |
| | | 50/50 | 0.65 (0.73-0.57) | 0.71 |
| | 1000 | 90/10 | 0.84 (0.91-0.77) | 0.73 |
| | | 80/20 | 0.82 (0.88-0.75) | 0.72 |
| | | 70/30 | 0.7 (0.77-0.62) | 0.62 |
| | | 60/40 | 0.65 (0.72-0.58) | 0.68 |
| | | 50/50 | 0.66 (0.73-0.58) | 0.73 |
| | 2000 | 90/10 | 0.88 (0.92-0.83) | 0.63 |
| | | 80/20 | 0.83 (0.88-0.78) | 0.74 |
| | | 70/30 | 0.76 (0.81-0.71) | 0.75 |
| | | 60/40 | 0.73 (0.79-0.68) | 0.77 |
| | | 50/50 | 0.66 (0.71-0.61) | 0.73 |
| | 3000 | 90/10 | 0.91 (0.94-0.88) | 0.72 |
| | | 80/20 | 0.81 (0.85-0.77) | 0.74 |
| | | 70/30 | 0.78 (0.81-0.74) | 0.77 |
| | | 60/40 | 0.68 (0.72-0.64) | 0.75 |
| | | 50/50 | 0.69 (0.73-0.65) | 0.76 |
| | 4000 | 90/10 | 0.89 (0.92-0.86) | 0.7 |
| | | 80/20 | 0.81 (0.85-0.77) | 0.75 |
| | | 70/30 | 0.78 (0.81-0.74) | 0.78 |
| | | 60/40 | 0.71 (0.75-0.67) | 0.77 |
| | | 50/50 | 0.69 (0.73-0.66) | 0.78 |
| | 5000 | 90/10 | 0.91 (0.93-0.89) | 0.72 |
| | | 80/20 | 0.82 (0.85-0.79) | 0.77 |
| | | 70/30 | 0.77 (0.8-0.74) | 0.74 |
| | | 60/40 | 0.74 (0.77-0.71) | 0.81 |
| | | 50/50 | 0.73 (0.76-0.69) | 0.81 |
| Naive Bayes | 200 | 90/10 | 0.91 (1-0.77) | 0.63 |
| | | 80/20 | 0.72 (0.91-0.52) | 0.51 |
| | | 70/30 | 0.73 (0.91-0.56) | 0.51 |
| | | 60/40 | 0.61 (0.79-0.41) | 0.76 |
| | | 50/50 | 0.62 (0.78-0.47) | 0.6 |
| | 400 | 90/10 | 0.88 (0.95-0.77) | 0.72 |
| | | 80/20 | 0.73 (0.85-0.6) | 0.54 |
| | | 70/30 | 0.73 (0.86-0.58) | 0.73 |
| | | 60/40 | 0.62 (0.76-0.46) | 0.54 |
| | | 50/50 | 0.65 (0.76-0.52) | 0.69 |
| | 500 | 90/10 | 0.94 (0.97-0.9) | 0.6 |
| | | 80/20 | 0.74 (0.84-0.62) | 0.73 |
| | | 70/30 | 0.73 (0.85-0.6) | 0.71 |
| | | 60/40 | 0.67 (0.8-0.54) | 0.72 |
| | | 50/50 | 0.68 (0.78-0.57) | 0.75 |
| | 600 | 90/10 | 0.8 (0.9-0.71) | 0.52 |
| | | 80/20 | 0.69 (0.8-0.57) | 0.68 |
| | | 70/30 | 0.72 (0.83-0.61) | 0.69 |
| | | 60/40 | 0.62 (0.74-0.51) | 0.67 |
| | | 50/50 | 0.64 (0.75-0.55) | 0.77 |
| | 800 | 90/10 | 0.86 (0.93-0.79) | 0.66 |
| | | 80/20 | 0.81 (0.89-0.72) | 0.7 |
| | | 70/30 | 0.77 (0.85-0.67) | 0.7 |
| | | 60/40 | 0.62 (0.72-0.52) | 0.76 |
| | | 50/50 | 0.72 (0.8-0.64) | 0.74 |
| | 1000 | 90/10 | 0.83 (0.9-0.75) | 0.74 |
| | | 80/20 | 0.77 (0.85-0.69) | 0.64 |
| | | 70/30 | 0.71 (0.79-0.62) | 0.61 |
| | | 60/40 | 0.59 (0.68-0.5) | 0.67 |
| | | 50/50 | 0.71 (0.77-0.63) | 0.72 |
| | 2000 | 90/10 | 0.85 (0.9-0.8) | 0.69 |
| | | 80/20 | 0.79 (0.85-0.74) | 0.75 |
| | | 70/30 | 0.76 (0.81-0.7) | 0.74 |
| | | 60/40 | 0.7 (0.76-0.63) | 0.75 |
| | | 50/50 | 0.62 (0.68-0.57) | 0.71 |
| | 3000 | 90/10 | 0.86 (0.9-0.82) | 0.7 |
| | | 80/20 | 0.78 (0.82-0.73) | 0.69 |
| | | 70/30 | 0.72 (0.77-0.67) | 0.71 |
| | | 60/40 | 0.64 (0.69-0.58) | 0.73 |
| | | 50/50 | 0.63 (0.68-0.59) | 0.71 |
| | 4000 | 90/10 | 0.84 (0.88-0.81) | 0.67 |
| | | 80/20 | 0.8 (0.85-0.76) | 0.74 |
| | | 70/30 | 0.73 (0.77-0.69) | 0.74 |
| | | 60/40 | 0.62 (0.66-0.57) | 0.71 |
| | | 50/50 | 0.66 (0.7-0.62) | 0.74 |
| | 5000 | 90/10 | 0.89 (0.92-0.86) | 0.67 |
| | | 80/20 | 0.81 (0.85-0.78) | 0.7 |
| | | 70/30 | 0.75 (0.78-0.71) | 0.74 |
| | | 60/40 | 0.72 (0.76-0.69) | 0.78 |
| | | 50/50 | 0.69 (0.73-0.66) | 0.78 |
| Random Forest | 200 | 90/10 | 0.91 (1-0.77) | 0.6 |
| | | 80/20 | 0.73 (0.91-0.56) | 0.5 |
| | | 70/30 | 0.71 (0.89-0.54) | 0.31 |
| | | 60/40 | 0.4 (0.61-0.2) | 0.71 |
| | | 50/50 | 0.35 (0.55-0.16) | 0.57 |
| | 400 | 90/10 | 0.89 (0.98-0.79) | 0.68 |
| | | 80/20 | 0.73 (0.85-0.58) | 0.61 |
| | | 70/30 | 0.67 (0.79-0.53) | 0.58 |
| | | 60/40 | 0.56 (0.72-0.4) | 0.56 |
| | | 50/50 | 0.65 (0.77-0.51) | 0.68 |
| | 500 | 90/10 | 0.93 (0.96-0.9) | 0.43 |
| | | 80/20 | 0.71 (0.82-0.58) | 0.68 |
| | | 70/30 | 0.71 (0.83-0.6) | 0.65 |
| | | 60/40 | 0.58 (0.72-0.46) | 0.7 |
| | | 50/50 | 0.63 (0.73-0.52) | 0.72 |
| | 600 | 90/10 | 0.8 (0.89-0.7) | 0.66 |
| | | 80/20 | 0.69 (0.79-0.57) | 0.71 |
| | | 70/30 | 0.62 (0.75-0.5) | 0.76 |
| | | 60/40 | 0.57 (0.7-0.44) | 0.64 |
| | | 50/50 | 0.64 (0.74-0.53) | 0.74 |
| | 800 | 90/10 | 0.86 (0.93-0.79) | 0.69 |
| | | 80/20 | 0.8 (0.88-0.7) | 0.61 |
| | | 70/30 | 0.71 (0.8-0.61) | 0.69 |
| | | 60/40 | 0.58 (0.68-0.47) | 0.76 |
| | | 50/50 | 0.7 (0.78-0.62) | 0.76 |
| | 1000 | 90/10 | 0.83 (0.89-0.75) | 0.68 |
| | | 80/20 | 0.74 (0.82-0.66) | 0.61 |
| | | 70/30 | 0.66 (0.75-0.56) | 0.63 |
| | | 60/40 | 0.58 (0.67-0.48) | 0.63 |
| | | 50/50 | 0.68 (0.76-0.61) | 0.7 |
| | 2000 | 90/10 | 0.84 (0.89-0.8) | 0.65 |
| | | 80/20 | 0.73 (0.79-0.67) | 0.71 |
| | | 70/30 | 0.72 (0.78-0.66) | 0.72 |
| | | 60/40 | 0.64 (0.7-0.58) | 0.74 |
| | | 50/50 | 0.61 (0.67-0.56) | 0.71 |
| | 3000 | 90/10 | 0.86 (0.9-0.82) | 0.71 |
| | | 80/20 | 0.75 (0.79-0.7) | 0.64 |
| | | 70/30 | 0.66 (0.72-0.6) | 0.71 |
| | | 60/40 | 0.59 (0.64-0.54) | 0.7 |
| | | 50/50 | 0.61 (0.66-0.56) | 0.67 |
| | 4000 | 90/10 | 0.84 (0.87-0.81) | 0.68 |
| | | 80/20 | 0.73 (0.77-0.69) | 0.72 |
| | | 70/30 | 0.65 (0.69-0.6) | 0.7 |
| | | 60/40 | 0.55 (0.6-0.5) | 0.68 |
| | | 50/50 | 0.64 (0.68-0.61) | 0.7 |
| | 5000 | 90/10 | 0.87 (0.9-0.84) | 0.72 |
| | | 80/20 | 0.76 (0.8-0.72) | 0.67 |
| | | 70/30 | 0.67 (0.71-0.63) | 0.71 |
| | | 60/40 | 0.63 (0.67-0.59) | 0.75 |
| | | 50/50 | 0.66 (0.69-0.63) | 0.75 |
| Stochiastic Gradient Descent | 200 | 90/10 | 0.91 (1-0.77) | 0.38 |
| | | 80/20 | 0.79 (0.93-0.6) | 0.52 |
| | | 70/30 | 0.65 (0.81-0.47) | 0.44 |
| | | 60/40 | 0.62 (0.78-0.43) | 0.69 |
| | | 50/50 | 0.59 (0.74-0.42) | 0.68 |
| | 400 | 90/10 | 0.87 (0.96-0.76) | 0.7 |
| | | 80/20 | 0.73 (0.86-0.59) | 0.62 |
| | | 70/30 | 0.73 (0.84-0.61) | 0.64 |
| | | 60/40 | 0.64 (0.77-0.52) | 0.59 |
| | | 50/50 | 0.62 (0.75-0.5) | 0.66 |
| | 500 | 90/10 | 0.92 (0.95-0.88) | 0.45 |
| | | 80/20 | 0.83 (0.91-0.74) | 0.71 |
| | | 70/30 | 0.72 (0.82-0.62) | 0.65 |
| | | 60/40 | 0.67 (0.77-0.56) | 0.69 |
| | | 50/50 | 0.64 (0.75-0.52) | 0.76 |
| | 600 | 90/10 | 0.8 (0.89-0.71) | 0.58 |
| | | 80/20 | 0.84 (0.92-0.74) | 0.65 |
| | | 70/30 | 0.74 (0.83-0.65) | 0.72 |
| | | 60/40 | 0.64 (0.74-0.53) | 0.66 |
| | | 50/50 | 0.64 (0.73-0.54) | 0.76 |
| | 800 | 90/10 | 0.9 (0.96-0.84) | 0.68 |
| | | 80/20 | 0.8 (0.88-0.72) | 0.63 |
| | | 70/30 | 0.76 (0.84-0.68) | 0.71 |
| | | 60/40 | 0.65 (0.72-0.57) | 0.72 |
| | | 50/50 | 0.58 (0.67-0.5) | 0.65 |
| | 1000 | 90/10 | 0.83 (0.9-0.76) | 0.69 |
| | | 80/20 | 0.83 (0.89-0.75) | 0.72 |
| | | 70/30 | 0.67 (0.75-0.59) | 0.61 |
| | | 60/40 | 0.67 (0.74-0.59) | 0.71 |
| | | 50/50 | 0.65 (0.72-0.58) | 0.7 |
| | 2000 | 90/10 | 0.88 (0.92-0.83) | 0.64 |
| | | 80/20 | 0.83 (0.88-0.78) | 0.74 |
| | | 70/30 | 0.76 (0.81-0.72) | 0.74 |
| | | 60/40 | 0.72 (0.77-0.67) | 0.75 |
| | | 50/50 | 0.64 (0.69-0.59) | 0.7 |
| | 3000 | 90/10 | 0.89 (0.92-0.85) | 0.7 |
| | | 80/20 | 0.8 (0.85-0.77) | 0.74 |
| | | 70/30 | 0.77 (0.81-0.73) | 0.76 |
| | | 60/40 | 0.68 (0.72-0.64) | 0.73 |
| | | 50/50 | 0.66 (0.7-0.62) | 0.74 |
| | 4000 | 90/10 | 0.88 (0.91-0.85) | 0.7 |
| | | 80/20 | 0.79 (0.82-0.75) | 0.73 |
| | | 70/30 | 0.76 (0.79-0.72) | 0.77 |
| | | 60/40 | 0.7 (0.74-0.67) | 0.76 |
| | | 50/50 | 0.69 (0.72-0.66) | 0.77 |
| | 5000 | 90/10 | 0.9 (0.92-0.87) | 0.72 |
| | | 80/20 | 0.82 (0.84-0.79) | 0.76 |
| | | 70/30 | 0.75 (0.78-0.72) | 0.72 |
| | | 60/40 | 0.74 (0.76-0.71) | 0.8 |
| | | 50/50 | 0.72 (0.75-0.69) | 0.81 |
| Support Vector Classifier | 200 | 90/10 | 0.91 (1-0.77) | 0.43 |
| | | 80/20 | 0.73 (0.91-0.52) | 0.49 |
| | | 70/30 | 0.73 (0.91-0.56) | 0.35 |
| | | 60/40 | 0.56 (0.76-0.36) | 0.78 |
| | | 50/50 | 0.48 (0.68-0.3) | 0.34 |
| | 400 | 90/10 | 0.89 (0.98-0.79) | 0.78 |
| | | 80/20 | 0.73 (0.86-0.58) | 0.59 |
| | | 70/30 | 0.73 (0.86-0.6) | 0.67 |
| | | 60/40 | 0.64 (0.79-0.47) | 0.58 |
| | | 50/50 | 0.65 (0.77-0.52) | 0.69 |
| | 500 | 90/10 | 0.93 (0.96-0.9) | 0.51 |
| | | 80/20 | 0.79 (0.89-0.67) | 0.75 |
| | | 70/30 | 0.75 (0.86-0.63) | 0.68 |
| | | 60/40 | 0.69 (0.8-0.57) | 0.73 |
| | | 50/50 | 0.67 (0.77-0.56) | 0.76 |
| | 600 | 90/10 | 0.8 (0.89-0.7) | 0.57 |
| | | 80/20 | 0.81 (0.9-0.69) | 0.7 |
| | | 70/30 | 0.76 (0.86-0.67) | 0.73 |
| | | 60/40 | 0.63 (0.74-0.51) | 0.67 |
| | | 50/50 | 0.63 (0.72-0.53) | 0.78 |
| | 800 | 90/10 | 0.88 (0.94-0.81) | 0.67 |
| | | 80/20 | 0.83 (0.9-0.74) | 0.66 |
| | | 70/30 | 0.77 (0.85-0.67) | 0.71 |
| | | 60/40 | 0.64 (0.73-0.55) | 0.78 |
| | | 50/50 | 0.72 (0.79-0.64) | 0.76 |
| | 1000 | 90/10 | 0.83 (0.9-0.76) | 0.75 |
| | | 80/20 | 0.79 (0.87-0.71) | 0.71 |
| | | 70/30 | 0.72 (0.8-0.64) | 0.6 |
| | | 60/40 | 0.59 (0.69-0.49) | 0.66 |
| | | 50/50 | 0.68 (0.75-0.6) | 0.73 |
| | 2000 | 90/10 | 0.87 (0.92-0.82) | 0.7 |
| | | 80/20 | 0.82 (0.87-0.77) | 0.77 |
| | | 70/30 | 0.78 (0.83-0.72) | 0.76 |
| | | 60/40 | 0.7 (0.75-0.64) | 0.77 |
| | | 50/50 | 0.66 (0.71-0.6) | 0.74 |
| | 3000 | 90/10 | 0.9 (0.93-0.86) | 0.74 |
| | | 80/20 | 0.79 (0.84-0.74) | 0.74 |
| | | 70/30 | 0.77 (0.81-0.72) | 0.77 |
| | | 60/40 | 0.69 (0.73-0.64) | 0.75 |
| | | 50/50 | 0.71 (0.76-0.67) | 0.77 |
| | 4000 | 90/10 | 0.89 (0.92-0.86) | 0.69 |
| | | 80/20 | 0.81 (0.85-0.77) | 0.75 |
| | | 70/30 | 0.78 (0.81-0.74) | 0.77 |
| | | 60/40 | 0.68 (0.71-0.64) | 0.78 |
| | | 50/50 | 0.7 (0.74-0.67) | 0.78 |
| | 5000 | 90/10 | 0.91 (0.93-0.88) | 0.72 |
| | | 80/20 | 0.83 (0.86-0.8) | 0.78 |
| | | 70/30 | 0.77 (0.81-0.74) | 0.76 |
| | | 60/40 | 0.76 (0.8-0.73) | 0.82 |
| | | 50/50 | 0.76 (0.79-0.73) | 0.83 |
| BERT | 200 | 90/10 | 0.86 (0.71-0.96) | 0.50 |
| | | 80/20 | 0.59 (0.42-0.75) | 0.50 |
| | | 70/30 | 0.55 (0.39-0.75) | 0.50 |
| | | 60/40 | 0.81 (0.67-0.95) | 0.77 |
| | | 50/50 | 0.53 (0.36-0.68) | 0.53 |
| | 400 | 90/10 | 0.8 (0.68-0.89) | 0.50 |
| | | 80/20 | 0.66 (0.53-0.8) | 0.50 |
| | | 70/30 | 0.53 (0.4-0.68) | 0.50 |
| | | 60/40 | 0.63 (0.51-0.74) | 0.63 |
| | | 50/50 | 0.65 (0.54-0.76) | 0.68 |
| | 500 | 90/10 | 0.85 (0.77-0.93) | 0.50 |
| | | 80/20 | 0.75 (0.65-0.85) | 0.60 |
| | | 70/30 | 0.71 (0.61-0.81) | 0.61 |
| | | 60/40 | 0.64 (0.54-0.73) | 0.65 |
| | | 50/50 | 0.58 (0.48-0.66) | 0.59 |
| | 600 | 90/10 | 0.84 (0.75-0.91) | 0.50 |
| | | 80/20 | 0.69 (0.58-0.79) | 0.50 |
| | | 70/30 | 0.62 (0.51-0.73) | 0.56 |
| | | 60/40 | 0.64 (0.55-0.72) | 0.63 |
| | | 50/50 | 0.62 (0.53-0.71) | 0.62 |
| | 800 | 90/10 | 0.84 (0.77-0.91) | 0.53 |
| | | 80/20 | 0.75 (0.66-0.83) | 0.62 |
| | | 70/30 | 0.69 (0.6-0.78) | 0.61 |
| | | 60/40 | 0.59 (0.49-0.68) | 0.59 |
| | | 50/50 | 0.67 (0.59-0.74) | 0.66 |
| | 1000 | 90/10 | 0.86 (0.8-0.91) | 0.50 |
| | | 80/20 | 0.74 (0.67-0.82) | 0.50 |
| | | 70/30 | 0.71 (0.64-0.77) | 0.60 |
| | | 60/40 | 0.61 (0.54-0.68) | 0.58 |
| | | 50/50 | 0.64 (0.58-0.71) | 0.65 |
| | 2000 | 90/10 | 0.9 (0.86-0.93) | 0.61 |
| | | 80/20 | 0.8 (0.76-0.84) | 0.65 |
| | | 70/30 | 0.72 (0.66-0.77) | 0.64 |
| | | 60/40 | 0.67 (0.63-0.72) | 0.65 |
| | | 50/50 | 0.63 (0.58-0.68) | 0.63 |
| | 3000 | 90/10 | 0.91 (0.88-0.94) | 0.66 |
| | | 80/20 | 0.8 (0.76-0.83) | 0.62 |
| | | 70/30 | 0.75 (0.71-0.79) | 0.67 |
| | | 60/40 | 0.67 (0.63-0.71) | 0.66 |
| | | 50/50 | 0.69 (0.65-0.73) | 0.69 |
| | 4000 | 90/10 | 0.91 (0.88-0.93) | 0.62 |
| | | 80/20 | 0.84 (0.81-0.87) | 0.66 |
| | | 70/30 | 0.77 (0.74-0.8) | 0.68 |
| | | 60/40 | 0.7 (0.67-0.74) | 0.67 |
| | | 50/50 | 0.7 (0.67-0.73) | 0.70 |
| | 5000 | 90/10 | 0.91 (0.89-0.93) | 0.64 |
| | | 80/20 | 0.83 (0.8-0.85) | 0.66 |
| | | 70/30 | 0.75 (0.72-0.77) | 0.68 |
| | | 60/40 | 0.69 (0.66-0.72) | 0.67 |
| | | 50/50 | 0.71 (0.68-0.73) | 0.71 |

**Appendix B**
Diagnosis codes within the Non-HTN group - delete: 2,127 words

HTN class includes ICD10 code 4019 i.e. Unspecified essential hypertension

Non-HTN class does not include any ICD10 codes beginning with 401x. The diagnoses included within this class are:

| | | |
|---|---|---|
| Shigella boydii | Pneumonia, organism NOS | Malignant neo colon NEC |
| Malign neopl prostate | Chr airway obstruct NEC | TB of limb bones-unspec |
| Cutaneous mycobacteria | Viral encephalitis NOS | TB limb bones-no exam |
| Strep sore throat | Postinflam pulm fibrosis | TB of bone NEC-unspec |
| Septicemia NOS | Erythema infectiosum | Malig neo pancreas NEC |
| Pneumococcus infect NOS | Hydronephrosis | TB of ureter-exam unkn |
| Subarachnoid hemorrhage | Trachoma NOS | TB of ureter-micro dx |
| Intracerebral hemorrhage | Early syph latent relaps | TB of ureter-histo dx |
| Subac scleros panenceph | TB lung infiltr-micro dx | Mal neo bronch/lung NEC |
| Bronchopneumonia org NOS | Malig neo tongue NOS | Malign neopl breast NEC |
| Sec malig neo lg bowel | TB of knee-unspec | Malig neo corpus uteri |
| Second malig neo liver | Benign neo skin leg | Malign neopl ovary |
| Sec malig neo urin NEC | Intramural leiomyoma | Malig neo bladder NEC |
| Sec mal neo brain/spine | Unc behav neo GI NEC | Mal neo parietal lobe |
| Secondary malig neo bone | Polycythemia vera | Mal neo cereb ventricle |
| Malignant neoplasm NOS | Hypothyroidism NOS | Malig neo brain NOS |
| Benign neoplasm lg bowel | Pancreatic disorder NEC | Mal neo lymph-head/neck |
| Ben neo liver/bile ducts | Neurohypophysis dis NEC | Mal neo lymph intra-abd |
| Benign neoplasm heart | Adrenal disorder NOS | Secondary malig neo lung |
| Benign neo skin eyelid | Testicular hypofunc NEC | Sec mal neo mediastinum |
| Pure hypercholesterolem | Protein-cal malnutr NOS | Second malig neo pleura |
| Pure hyperglyceridemia | Anemia NOS | Sec malig neo sm bowel |
| Hyperlipidemia NEC/NOS | Thrombocytopenia NOS | Neuropathy in diabetes |
| Lipoid metabol dis NOS | Wbc disease NEC | Glaucoma NOS |
| Gout NOS | Delirium d/t other cond | E coli septicemia |
| Hyperosmolality | Transient mental dis NOS | Hearing loss NOS |
| Hyposmolality | Mental disor NEC oth dis | Mitral insuf/aort stenos |
| Acidosis | Bipolor I current NOS | Mitral/aortic val insuff |
| Alkalosis | Obsessive-compulsive dis | Mitr/aortic mult involv |
| Hyperpotassemia | Dysthymic disorder | Tricuspid valve disease |
| Hypopotassemia | Nonpsychotic disord NOS | Mth sus Stph aur els/NOS |
| Chr blood loss anemia | Tobacco use disorder | Angina pectoris NEC/NOS |
| Iron defic anemia NOS | Bacterial meningitis NOS | Chr ischemic hrt dis NEC |
| B12 defic anemia NEC | Obstructiv hydrocephalus | Chr pulmon heart dis NEC |
| Ac posthemorrhag anemia | Paralysis agitans | Periapical abscess |
| Helicobacter pylori | Grand mal status | Sialoadenitis |
| Pericardial disease NOS | Compression of brain | Achalasia & cardiospasm |
| Mitral valve disorder | Trigeminal neuralgia | Esophageal stricture |
| Aortic valve disorder | Rupt abd aortic aneurysm | Mallory-weiss syndrome |
| Prim cardiomyopathy NEC | Abdom aortic aneurysm | Acq pyloric stenosis |
| Atriovent block complete | Thracabd anurysm wo rupt | Hernia, site NEC w obstr |
| Parox atrial tachycardia | Periph vascular dis NOS | Umbilical hernia |
| Parox ventric tachycard | Orthostatic hypotension | Diaphragmatic hernia |
| Cardiac arrest | Hypotension NOS | Reg enterit sm/lg intest |
| CHF NOS | Acute uri NOS | Ulceratve colitis unspcf |
| Cardiomegaly | Acute bronchitis | Allrgic gastro & colitis |
| Heart disease NOS | Chronic sinusitis NOS | Noninf gastroenterit NEC |
| Subdural hemorrhage | Vocal cord/larynx polyp | Rubella encephalitis |
| Intracranial hemorr NOS | Emphysema NEC | Peritoneal adhesions |
| Trans cereb ischemia NOS | Food/vomit pneumonitis | Rectal prolapse |
| Nonrupt cerebral aneurym | Pleural effusion NOS | Alcohol cirrhosis liver |
| Cerebrovasc disease NEC | Iatrogenic pneumothorax | Cirrhosis of liver NOS |
| Ruptur thoracic aneurysm | Abscess of lung | Chronic liver dis NEC |
| Thoracic aortic aneurysm | Pulmonary collapse | Hepatitis NOS |
| Blood in stool | Acute lung edema NOS | Cholangitis |
| Gastrointest hemorr NOS | Cervicalgia | Dis of biliary tract NEC |
| Human herpesvr encph NEC | Sciatica | Chronic pancreatitis |
| Ac kidny fail, tubr necr | Backache NOS | Pancreat cyst/pseudocyst |
| Acute kidney failure NOS | Other back symptoms | Hematemesis |
| End stage renal disease | Myalgia and myositis NOS | Patent ductus arteriosus |
| Chronic kidney dis NOS | Brain anomaly NEC | Intestinal anomaly NEC |
| Yatapoxvirus infectn NOS | Accessory auricle | Bladder exstrophy |
| Tanapox | Tetralogy of fallot | Down's syndrome |
| Stricture of ureter | Ventricular sept defect | Gonadal dysgenesis |
| Renal & ureteral dis NOS | Secundum atrial sept def | Hamartoses NEC |
| Urin tract infection NOS | Septal closure anom NEC | Congenital anomaly NOS |
| Noninfl dis ova/adnx NEC | Cong tricusp atres/sten | Abn plac NEC/NOS aff NB |
| Excessive menstruation | Cong aorta valv insuffic | Oth umbil cord compress |
| Cellulitis of foot | NB integument cond NEC | Exceptionally large baby |
| Pilonidal cyst w/o absc | Syncope and collapse | Heavy-for-date infan NEC |
| Diaper or napkin rash | Headache | Fetal distrs dur lab/del |
| Lupus erythematosus | Aphasia | NB transitory tachypnea |
| Other psoriasis | Epistaxis | NB cutaneous hemorrhage |
| Hpt C acute wo hpat coma | Tachycardia NOS | NB hemolyt dis-abo isoim |
| Chrnc hpt C wo hpat coma | Cardiac murmurs NEC | Neonat jaund preterm del |
| Hpt C w/o hepat coma NOS | Resp sys/chest symp NEC | Fetal/neonatal jaund NOS |
| Skin disorders NEC | Oliguria & anuria | Infant diabet mother syn |
| Sicca syndrome | Abn blood chemistry NEC | Neonatal dehydration |
| Rheumatoid arthritis | Abn find-stool contents | Neonatal hypoglycemia |
| Cerv spondyl w myelopath | Debility NOS | Perinatal intest perfor |
| Acute syphil meningitis | Fx malar/maxillary-close | NB hypothermia NEC |
| Contusion of chest wall | Fx malar/maxillary-open | Congenital hydrocele |
| Foreign body esophagus | Fx orbital floor-closed | Ch myl leuk wo achv rmsn |
| Injury femoral nerve | Fx orbital floor-open | Hemangioma skin |
| Pois-arom analgesics NEC | Fx facial bone NEC-close | Hemangioma NEC |
| Pois-anticonvul NEC/NOS | Fx lumbar vertebra-close | Myelodysplastic synd NOS |
| Pois-benzodiazepine tran | Fracture of sternum-clos | Tox dif goiter no crisis |
| Toxic eff ethyl alcohol | Fracture acetabulum-clos | DMII wo cmp nt st uncntr |
| Oth VD chlm trch unsp st | Traum pneumothorax-close | DMI wo cmp nt st uncntrl |
| Surg compl-heart | Traum pneumothorax-open | DMII wo cmp uncntrld |
| Accidental op laceration | Lac eyelid inv lacrm pas | DMII keto nt st uncntrld |
| Vasc comp med care NEC | Open wound of scalp | DMI keto nt st uncntrld |
| Second malig neo genital | Open wound of chest | DMII ketoacd uncontrold |
| Hdgk dis unsp xtrndl org | Open wound hand w tendon | DMI ketoacd uncontrold |
| Mycs fng unsp xtrndl org | Amputation finger | DMII renl nt st uncntrld |
| Mult mye w/o achv rmson | Abrasion head | DMII neuro nt st uncntrl |
| Cardiac dysrhythmias NEC | Abrasion forearm | DMII oth nt st uncntrld |
| Diastolc hrt failure NOS | Ulcer of heel & midfoot | DMI oth nt st uncntrld |
| Chr diastolic hrt fail | Osteoarthros NOS-unspec | Acute gouty arthropathy |
| Ill-defined hrt dis NEC | Joint symptom NEC-pelvis | Dehydration |
| Ocl crtd art wo infrct | Necrotizing fasciitis | Obesity NOS |
| Ocl mlt bi art wo infrct | Rhabdomyolysis | Morbid obesity |
| Crbl emblsm w infrct | Osteoporosis NOS | Anemia in neoplastic dis |
| Crbl art ocl NOS w infrc | Malunion of fracture | Alcohol withdrawal |
| Late eff CV dis-aphasia | Bone & cartilage dis NOS | Dementia w/o behav dist |
| Late ef-spch/lang df NEC | Forearm deformity NOS | Paranoid schizo-unspec |
| Late ef-hemplga side NOS | Kyphosis NOS | Schizoaffective dis NOS |
| Late effect CV dis NEC | Thoracogenic scoliosis | Schizoafftv dis-chr/exac |
| Dsct of thoracic aorta | Spin bif w hydrceph-cerv | Schizophrenia NOS-unspec |
| Upper extremity embolism | Spec lacrimal pass anom | Bipol I currnt manic NOS |
| Bleed esoph var oth dis | Ex ear anm NEC-impr hear | Anxiety state NOS |
| Iatrogenc hypotnsion NEC | Ostium primum defect | Conversion disorder |
| Obs chr bronc w(ac) exac | Cong pulmon valve stenos | Borderline personality |
| Ext asthma w status asth | Cong heart anomaly NEC | Ac alcohol intox-contin |
| Chronic obst asthma NOS | Great vein anomaly NEC | Alcoh dep NEC/NOS-unspec |
| Asthma NOS | Cerebrovascular anomaly | Alcoh dep NEC/NOS-contin |
| Asthma NOS w (ac) exac | Spinal vessel anomaly | Opioid dependence-unspec |
| Acute respiratry failure | Biliary & liver anom NEC | Opioid dependence-contin |
| Other pulmonary insuff | Hypospadias | Drug depend NOS-unspec |
| Acute & chronc resp fail | Obst def ren plv&urt NEC | Alcohol abuse-unspec |
| Tracheostomy - mech comp | Congn anoml abd wall NEC | Alcohol abuse-continuous |
| Other esophagitis | Cong skin pigment anomal | Cocaine abuse-unspec |
| Ulc esophagus w/o bleed | Nox sub NEC aff NB/fetus | Amphetamine abuse-unspec |
| Esophageal reflux | Lt-for-dates 1750-1999g | Drug abuse NEC-unspec |
| Barrett's esophagus | Lt-for-dates 2000-2499g | Attn deficit w hyperact |
| Chr stomach ulc w hem | Preterm NEC 1750-1999g | Obstructive sleep apnea |
| Stomach ulcer NOS | Preterm NEC 2000-2499g | Dementia w Lewy bodies |
| Chr duoden ulcer w hem | Preterm NEC 2500+g | Amyotrophic sclerosis |
| Duodenal ulcer NOS | 33-34 comp wks gestation | Psymotr epil w/o int epi |
| Chr marginal ulc w perf | 35-36 comp wks gestation | Part epil w/o intr epil |
| Oth spf gastrt w hmrhg | 37+ comp wks gestation | Epilep NOS w/o intr epil |
| Gstr/ddnts NOS w/o hmrhg | Injuries to scalp NEC | Othr migrne wo ntrc mgrn |
| Gstr/ddnts NOS w hmrhg | Primary apnea of newborn | Migrne unsp wo ntrc mgrn |
| Duodenitis w/o hmrhg | Cyanotic attack, newborn | Rheumatic heart failure |
| Gastroduodenal dis NEC | Resp failure of newborn | Hy kid NOS w cr kid I-IV |
| Intestinal obstruct NEC | Resp prob after brth NEC | Hyp kid NOS w cr kid V |
| Dvrtcli colon w/o hmrhg | Bacteremia of newborn | AMI anterior wall, init |
| Dvrtclo colon w hmrhg | Neonatal tachycardia | AMI inferolateral, init |
| Constipation NOS | Meconium staining | AMI inferopost, initial |
| Peritonitis (acute) gen | Perinatal condition NEC | AMI inferior wall, init |
| Peritonitis NEC | Other alter consciousnes | True post infarct, init |
| Ulceration of intestine | Convulsions NEC | Subendo infarct, initial |
| Perforation of intestine | Sleep apnea NOS | Subendo infarct, subseq |
| Angio intes w hmrhg | Cardiogenic shock | AMI NEC, initial |
| Cholelith w ac cholecyst | Septic shock | AMI NOS, initial |
| Cholelithiasis NOS | Shock w/o trauma NEC | AMI NOS, subsequent |
| Gall&bil cal w/oth w obs | Chest pain NOS | Cor ath unsp vsl ntv/gft |
| Nephritis NOS in oth dis | Nausea with vomiting | Crnry athrscl natve vssl |
| Ac pyelonephritis NOS | Diarrhea | Crn ath atlg vn bps grft |
| Neurogenic bladder NOS | Retention urine NOS | Cor ath artry bypas grft |
| BPH w/o urinary obs/LUTS | Urinary frequency | Aneurysm coronary vessel |
| Legal abort w hemorr-inc | Drop, hematocrit, precip | Atrioven block-mobitz ii |
| Mild/NOS preeclamp-p/p | Abnrml coagultion prfile | Conduction disorder NEC |
| Anemia-delivered w p/p | Hypoxemia | Atrial fibrillation |
| CV dis NEC-antepartum | Cl skl vlt fx/cerebr lac | Atrial flutter |
| Postpa hem NEC-del w p/p | Cl skl base fx/cereb lac | Ventricular fibrillation |
| P/p coag def-del w p/p | Cl skl base fx/menin hem | Sinoatrial node dysfunct |
| Peripartum card-postpart | Cl skul base fx w/o coma | Abn react-anastom/graft |
| Brain lacer NEC w/o coma | Cl skl fx NEC/mening hem | Abn reac-organ rem NEC |
| Subarach hem-brief coma | Cl skul fx NEC-deep coma | Abn react-surg proc NEC |
| Subarach hem-deep coma | Cl skl w oth fx-coma NOS | Abn react-cardiac cath |
| Subarach hem-coma NOS | Fx c2 vertebra-closed | Abn react-radiotherapy |
| Traumatic subdural hem | Fx mult cervical vert-cl | Abn react-procedure NOS |
| Subdural hem w/o coma | C5-c7 fx-cl/ant cord syn | Fall on stair/step NEC |
| Traumatic brain hem NEC | Fracture three ribs-clos | Fall from ladder |
| Brain hem NEC-coma NOS | Fracture seven ribs-clos | Diving accident |
| Heart contusion-closed | Fx tibia NOS-closed | Fall-1 level to oth NEC |
| Lung contusion-closed | Fx tibia w fibula NOS-cl | Fall from slipping NEC |
| Duodenum injury-closed | Disloc 2nd cerv vert-cl | Fall NOS |
| Sigmoid colon inj-closed | Sprain of ankle NOS | Resp obstr-food inhal |
| Liver hematoma/contusion | Brain laceration NEC | FB entering oth orifice |
| Liver lacerat unspcf cls | Comp-oth vasc dev/graft | Struck by falling object |
| Liver injury NEC-closed | Hemorrhage complic proc | Woodworking machine acc |
| Spleen injury NOS-closed | Hematoma complic proc | Machinery accident NEC |
| Spleen hematoma-closed | Other postop infection | Acc-cutting instrum NEC |
| Spleen disruption-clos | Non-healing surgcl wound | Hunting rifle accident |
| Spleen injury NEC-closed | Oth spcf cmplc procd NEC | Adv eff cephalosporin |
| Open wound of forehead | Mv collision NOS-driver | Adv eff antineoplastic |
| Open wound of jaw | Mv collision NOS-pasngr | Adv eff opiates |
| Open wound of face NEC | Mv-oth veh coll-driver | Adv eff analgesic NOS |
| Poisoning-opiates NEC | Mv coll w ped-ped cycl | Adv eff sedat/hypnot NEC |
| Pois-propionic acid derv | Mv coll w pedest-pedest | Adv eff coronary vasodil |
| Severe sepsis | Loss control mv acc-driv | Poison-analgesics |
| SIRS-noninf w/o ac or ds | Loss control mv acc-psgr | Poison-drug/medicin NEC |
| Malfunc prosth hrt valve | Traffic acc NOS-driver | Poison-solid/liquid NEC |
| Periprosthetc osteolysis | Ped cycl acc-ped cyclist | Unarmed fight or brawl |
| React-int pros devic NEC | Accid in recreation area | Assault-cutting instr |
| Comp-heart valve prosth | Acc poisn-benzdiaz tranq | Assault-striking w obj |
| Comp-oth cardiac device | Abn react-artif implant | Assault NOS |
| Undeter pois-sol/liq NEC | Prsnl hst colonic polyps | Undeterm pois-analgesics |
| Need prphyl vc vrl hepat | Prsnl hst ot spf dgst ds | Undeterm pois-psychotrop |
| Asymp hiv infectn status | Trnspl status-pancreas | Heart valve replac NEC |
| Hx of colonic malignancy | History of tobacco use | Joint replaced hip |
| Hx-bronchogenic malignan | Family hx-breast malig | Joint replaced knee |
| Hx of breast malignancy | Fam hx-ischem heart dis | Status cardiac pacemaker |
| Hx-uterus malignancy NEC | Fam hx-diabetes mellitus | Acq absnce breast/nipple |
| Hx-prostatic malignancy | NB obsrv suspct infect | Acquired absence kidney |
| Hx of bladder malignancy | NB obs genetc/metabl cnd | Acq absence of lung |
| Hx-lymphatic malign NEC | NB obsrv oth suspct cond | Aortocoronary bypass |
| Hx-malig skin melanoma | Single lb in-hosp w/o cs | Status-post ptca |
| Hx-skin malignancy NEC | Single lb in-hosp w cs | Routine circumcision |
| Hx of brain malignancy | Singl livebrn-before adm | Long-term use anticoagul |
| Hx of affective disorder | Twin-mate lb-hosp w/o cs | Long-term use of insulin |
| Personal histry malaria | Twin-mate lb-in hos w cs | Wait adm to oth facility |
| Hx-ven thrombosis/embols | Twin-mate sb-hosp w cs | No proc/contraindication |
| Hx-circulatory dis NEC | Oth mult lb-in hosp w cs | No proc/patient decision |
| Observ-accident NEC | Kidney transplant status | Exam-clincal trial |

**Appendix C - Diabetes**

All classifiers, sample sizes, class proportions and scores

| Classifier | Sample Size | Class Proportion (class 1/ class 0) | F1-score Weighted avg. (95% CI) | AUC Score |
|---|---|---|---|---|
| Logistic Regression | 200 | 90/10 | 0.82 (0.95-0.64) | 0.71 |
| | | 80/20 | 0.77 (0.91-0.6) | 0.83 |
| | | 70/30 | 0.83 (0.97-0.67) | 0.85 |
| | | 60/40 | 0.67 (0.83-0.48) | 0.84 |
| | | 50/50 | 0.75 (0.88-0.61) | 0.78 |
| | 400 | 90/10 | 0.95 (1-0.88) | 0.96 |
| | | 80/20 | 0.87 (0.95-0.77) | 0.87 |
| | | 70/30 | 0.73 (0.85-0.61) | 0.78 |
| | | 60/40 | 0.69 (0.81-0.57) | 0.75 |
| | | 50/50 | 0.73 (0.84-0.61) | 0.85 |
| | 500 | 90/10 | 0.84 (0.94-0.74) | 0.71 |
| | | 80/20 | 0.83 (0.92-0.73) | 0.82 |
| | | 70/30 | 0.74 (0.84-0.63) | 0.85 |
| | | 60/40 | 0.76 (0.85-0.66) | 0.86 |
| | | 50/50 | 0.71 (0.81-0.6) | 0.82 |
| | 600 | 90/10 | 0.9 (0.97-0.82) | 0.75 |
| | | 80/20 | 0.75 (0.86-0.64) | 0.71 |
| | | 70/30 | 0.72 (0.81-0.61) | 0.78 |
| | | 60/40 | 0.72 (0.8-0.63) | 0.8 |
| | | 50/50 | 0.73 (0.82-0.64) | 0.83 |
| | 800 | 90/10 | 0.9 (0.95-0.83) | 0.86 |
| | | 80/20 | 0.83 (0.91-0.75) | 0.77 |
| | | 70/30 | 0.75 (0.83-0.66) | 0.79 |
| | | 60/40 | 0.72 (0.79-0.65) | 0.81 |
| | | 50/50 | 0.69 (0.76-0.61) | 0.78 |
| | 1000 | 90/10 | 0.88 (0.93-0.82) | 0.84 |
| | | 80/20 | 0.76 (0.84-0.69) | 0.8 |
| | | 70/30 | 0.79 (0.85-0.73) | 0.85 |
| | | 60/40 | 0.75 (0.81-0.67) | 0.84 |
| | | 50/50 | 0.74 (0.81-0.67) | 0.84 |
| | 2000 | 90/10 | 0.89 (0.93-0.84) | 0.8 |
| | | 80/20 | 0.86 (0.9-0.82) | 0.84 |
| | | 70/30 | 0.79 (0.84-0.75) | 0.86 |
| | | 60/40 | 0.81 (0.85-0.76) | 0.88 |
| | | 50/50 | 0.76 (0.8-0.71) | 0.86 |
| | 3000 | 90/10 | 0.92 (0.95-0.9) | 0.88 |
| | | 80/20 | 0.87 (0.9-0.83) | 0.85 |
| | | 70/30 | 0.83 (0.86-0.8) | 0.88 |
| | | 60/40 | 0.8 (0.84-0.77) | 0.89 |
| | | 50/50 | 0.8 (0.84-0.77) | 0.88 |
| | 4000 | 90/10 | 0.93 (0.95-0.9) | 0.86 |
| | | 80/20 | 0.87 (0.9-0.84) | 0.89 |
| | | 70/30 | 0.86 (0.89-0.83) | 0.91 |
| | | 60/40 | 0.82 (0.85-0.79) | 0.9 |
| | | 50/50 | 0.8 (0.83-0.77) | 0.89 |
| | 5000 | 90/10 | 0.92 (0.94-0.9) | 0.87 |
| | | 80/20 | 0.85 (0.88-0.83) | 0.88 |
| | | 70/30 | 0.86 (0.88-0.83) | 0.9 |
| | | 60/40 | 0.83 (0.86-0.81) | 0.92 |
| | | 50/50 | 0.85 (0.87-0.82) | 0.92 |
| Decision Tree | 200 | 90/10 | 0.8 (0.95-0.63) | 0.48 |
| | | 80/20 | 0.82 (0.94-0.7) | 0.76 |
| | | 70/30 | 0.73 (0.86-0.58) | 0.7 |
| | | 60/40 | 0.69 (0.84-0.53) | 0.69 |
| | | 50/50 | 0.47 (0.66-0.31) | 0.47 |
| | 400 | 90/10 | 0.93 (0.98-0.86) | 0.48 |
| | | 80/20 | 0.78 (0.87-0.68) | 0.66 |
| | | 70/30 | 0.68 (0.8-0.56) | 0.53 |
| | | 60/40 | 0.76 (0.86-0.64) | 0.73 |
| | | 50/50 | 0.71 (0.83-0.59) | 0.71 |
| | 500 | 90/10 | 0.85 (0.93-0.76) | 0.61 |
| | | 80/20 | 0.83 (0.91-0.74) | 0.69 |
| | | 70/30 | 0.76 (0.85-0.66) | 0.73 |
| | | 60/40 | 0.78 (0.86-0.69) | 0.78 |
| | | 50/50 | 0.84 (0.91-0.75) | 0.84 |
| | 600 | 90/10 | 0.89 (0.96-0.82) | 0.64 |
| | | 80/20 | 0.68 (0.79-0.58) | 0.51 |
| | | 70/30 | 0.66 (0.76-0.56) | 0.61 |
| | | 60/40 | 0.63 (0.71-0.53) | 0.64 |
| | | 50/50 | 0.7 (0.78-0.61) | 0.71 |
| | 800 | 90/10 | 0.89 (0.95-0.84) | 0.72 |
| | | 80/20 | 0.78 (0.85-0.7) | 0.61 |
| | | 70/30 | 0.75 (0.82-0.66) | 0.69 |
| | | 60/40 | 0.67 (0.76-0.59) | 0.66 |
| | | 50/50 | 0.68 (0.75-0.6) | 0.69 |
| | 1000 | 90/10 | 0.86 (0.91-0.81) | 0.69 |
| | | 80/20 | 0.72 (0.79-0.64) | 0.6 |
| | | 70/30 | 0.8 (0.86-0.73) | 0.73 |
| | | 60/40 | 0.7 (0.76-0.63) | 0.7 |
| | | 50/50 | 0.61 (0.69-0.55) | 0.61 |
| | 2000 | 90/10 | 0.88 (0.91-0.84) | 0.6 |
| | | 80/20 | 0.85 (0.88-0.81) | 0.75 |
| | | 70/30 | 0.78 (0.83-0.74) | 0.74 |
| | | 60/40 | 0.8 (0.83-0.75) | 0.8 |
| | | 50/50 | 0.76 (0.81-0.72) | 0.77 |
| | 3000 | 90/10 | 0.92 (0.95-0.89) | 0.79 |
| | | 80/20 | 0.83 (0.86-0.8) | 0.76 |
| | | 70/30 | 0.77 (0.8-0.74) | 0.74 |
| | | 60/40 | 0.82 (0.85-0.78) | 0.83 |
| | | 50/50 | 0.8 (0.83-0.76) | 0.8 |
| | 4000 | 90/10 | 0.9 (0.92-0.88) | 0.75 |
| | | 80/20 | 0.86 (0.89-0.83) | 0.79 |
| | | 70/30 | 0.85 (0.87-0.82) | 0.84 |
| | | 60/40 | 0.83 (0.86-0.8) | 0.83 |
| | | 50/50 | 0.82 (0.85-0.79) | 0.82 |
| | 5000 | 90/10 | 0.9 (0.92-0.88) | 0.76 |
| | | 80/20 | 0.87 (0.89-0.84) | 0.8 |
| | | 70/30 | 0.87 (0.89-0.84) | 0.86 |
| | | 60/40 | 0.85 (0.87-0.83) | 0.85 |
| | | 50/50 | 0.87 (0.9-0.85) | 0.87 |
| K-Nearest Neighbour | 200 | 90/10 | 0.82 (0.95-0.6) | 0.45 |
| | | 80/20 | 0.86 (0.97-0.71) | 0.81 |
| | | 70/30 | 0.81 (0.94-0.66) | 0.82 |
| | | 60/40 | 0.69 (0.84-0.53) | 0.76 |
| | | 50/50 | 0.58 (0.75-0.41) | 0.71 |
| | 400 | 90/10 | 0.97 (1-0.92) | 0.94 |
| | | 80/20 | 0.86 (0.94-0.78) | 0.88 |
| | | 70/30 | 0.69 (0.8-0.57) | 0.73 |
| | | 60/40 | 0.71 (0.82-0.59) | 0.72 |
| | | 50/50 | 0.68 (0.8-0.56) | 0.79 |
| | 500 | 90/10 | 0.84 (0.92-0.75) | 0.67 |
| | | 80/20 | 0.87 (0.94-0.79) | 0.76 |
| | | 70/30 | 0.72 (0.82-0.6) | 0.81 |
| | | 60/40 | 0.73 (0.83-0.63) | 0.81 |
| | | 50/50 | 0.69 (0.79-0.59) | 0.77 |
| | 600 | 90/10 | 0.89 (0.96-0.81) | 0.65 |
| | | 80/20 | 0.76 (0.84-0.66) | 0.74 |
| | | 70/30 | 0.77 (0.86-0.68) | 0.75 |
| | | 60/40 | 0.68 (0.78-0.59) | 0.75 |
| | | 50/50 | 0.68 (0.77-0.58) | 0.77 |
| | 800 | 90/10 | 0.92 (0.97-0.85) | 0.79 |
| | | 80/20 | 0.81 (0.88-0.74) | 0.78 |
| | | 70/30 | 0.7 (0.79-0.61) | 0.74 |
| | | 60/40 | 0.71 (0.78-0.64) | 0.8 |
| | | 50/50 | 0.62 (0.7-0.53) | 0.7 |
| | 1000 | 90/10 | 0.87 (0.92-0.81) | 0.73 |
| | | 80/20 | 0.76 (0.84-0.68) | 0.74 |
| | | 70/30 | 0.81 (0.87-0.75) | 0.83 |
| | | 60/40 | 0.77 (0.83-0.7) | 0.82 |
| | | 50/50 | 0.7 (0.77-0.63) | 0.77 |
| | 2000 | 90/10 | 0.88 (0.92-0.84) | 0.68 |
| | | 80/20 | 0.79 (0.84-0.74) | 0.73 |
| | | 70/30 | 0.74 (0.79-0.68) | 0.81 |
| | | 60/40 | 0.73 (0.77-0.68) | 0.81 |
| | | 50/50 | 0.71 (0.77-0.66) | 0.79 |
| | 3000 | 90/10 | 0.9 (0.92-0.87) | 0.8 |
| | | 80/20 | 0.81 (0.85-0.77) | 0.82 |
| | | 70/30 | 0.77 (0.8-0.73) | 0.81 |
| | | 60/40 | 0.77 (0.81-0.73) | 0.85 |
| | | 50/50 | 0.73 (0.77-0.69) | 0.83 |
| | 4000 | 90/10 | 0.92 (0.94-0.89) | 0.77 |
| | | 80/20 | 0.82 (0.85-0.79) | 0.82 |
| | | 70/30 | 0.78 (0.82-0.75) | 0.82 |
| | | 60/40 | 0.77 (0.8-0.74) | 0.85 |
| | | 50/50 | 0.76 (0.8-0.73) | 0.84 |
| | 5000 | 90/10 | 0.91 (0.93-0.88) | 0.8 |
| | | 80/20 | 0.85 (0.88-0.82) | 0.82 |
| | | 70/30 | 0.8 (0.83-0.77) | 0.85 |
| | | 60/40 | 0.78 (0.8-0.75) | 0.87 |
| | | 50/50 | 0.76 (0.79-0.73) | 0.85 |
| Linear Support Vector Classifier | 200 | 90/10 | 0.82 (0.95-0.64) | 0.67 |
| | | 80/20 | 0.81 (0.94-0.67) | 0.81 |
| | | 70/30 | 0.84 (0.97-0.69) | 0.83 |
| | | 60/40 | 0.65 (0.81-0.47) | 0.84 |
| | | 50/50 | 0.75 (0.91-0.61) | 0.78 |
| | 400 | 90/10 | 0.95 (1-0.88) | 0.96 |
| | | 80/20 | 0.92 (0.98-0.84) | 0.88 |
| | | 70/30 | 0.71 (0.84-0.58) | 0.79 |
| | | 60/40 | 0.66 (0.77-0.55) | 0.75 |
| | | 50/50 | 0.71 (0.82-0.58) | 0.85 |
| | 500 | 90/10 | 0.87 (0.95-0.78) | 0.69 |
| | | 80/20 | 0.85 (0.93-0.76) | 0.79 |
| | | 70/30 | 0.74 (0.84-0.64) | 0.86 |
| | | 60/40 | 0.76 (0.85-0.66) | 0.86 |
| | | 50/50 | 0.74 (0.83-0.64) | 0.82 |
| | 600 | 90/10 | 0.91 (0.97-0.82) | 0.69 |
| | | 80/20 | 0.75 (0.85-0.63) | 0.7 |
| | | 70/30 | 0.71 (0.8-0.61) | 0.78 |
| | | 60/40 | 0.7 (0.78-0.6) | 0.79 |
| | | 50/50 | 0.73 (0.8-0.64) | 0.82 |
| | 800 | 90/10 | 0.89 (0.95-0.82) | 0.89 |
| | | 80/20 | 0.86 (0.92-0.78) | 0.73 |
| | | 70/30 | 0.75 (0.83-0.65) | 0.78 |
| | | 60/40 | 0.72 (0.79-0.63) | 0.8 |
| | | 50/50 | 0.7 (0.78-0.62) | 0.78 |
| | 1000 | 90/10 | 0.88 (0.94-0.82) | 0.8 |
| | | 80/20 | 0.76 (0.83-0.67) | 0.77 |
| | | 70/30 | 0.81 (0.87-0.74) | 0.84 |
| | | 60/40 | 0.76 (0.83-0.7) | 0.84 |
| | | 50/50 | 0.74 (0.81-0.66) | 0.85 |
| | 2000 | 90/10 | 0.89 (0.93-0.84) | 0.75 |
| | | 80/20 | 0.86 (0.9-0.82) | 0.83 |
| | | 70/30 | 0.81 (0.85-0.76) | 0.85 |
| | | 60/40 | 0.81 (0.85-0.76) | 0.88 |
| | | 50/50 | 0.76 (0.81-0.72) | 0.87 |
| | 3000 | 90/10 | 0.93 (0.96-0.9) | 0.86 |
| | | 80/20 | 0.86 (0.89-0.82) | 0.84 |
| | | 70/30 | 0.83 (0.86-0.8) | 0.87 |
| | | 60/40 | 0.8 (0.84-0.77) | 0.89 |
| | | 50/50 | 0.8 (0.84-0.77) | 0.88 |
| | 4000 | 90/10 | 0.93 (0.96-0.91) | 0.85 |
| | | 80/20 | 0.87 (0.9-0.84) | 0.88 |
| | | 70/30 | 0.86 (0.88-0.83) | 0.91 |
| | | 60/40 | 0.83 (0.86-0.8) | 0.9 |
| | | 50/50 | 0.81 (0.84-0.78) | 0.89 |
| | 5000 | 90/10 | 0.93 (0.95-0.9) | 0.84 |
| | | 80/20 | 0.86 (0.88-0.83) | 0.87 |
| | | 70/30 | 0.86 (0.88-0.83) | 0.89 |
| | | 60/40 | 0.82 (0.85-0.8) | 0.91 |
| | | 50/50 | 0.83 (0.86-0.81) | 0.92 |
| Naive Bayes | 200 | 90/10 | 0.82 (0.95-0.64) | 0.79 |
| | | 80/20 | 0.73 (0.91-0.56) | 0.81 |
| | | 70/30 | 0.67 (0.85-0.46) | 0.8 |
| | | 60/40 | 0.48 (0.67-0.27) | 0.82 |
| | | 50/50 | 0.66 (0.81-0.48) | 0.78 |
| | 400 | 90/10 | 0.95 (1-0.88) | 0.91 |
| | | 80/20 | 0.83 (0.94-0.71) | 0.84 |
| | | 70/30 | 0.61 (0.75-0.49) | 0.75 |
| | | 60/40 | 0.73 (0.86-0.62) | 0.81 |
| | | 50/50 | 0.62 (0.74-0.5) | 0.79 |
| | 500 | 90/10 | 0.82 (0.93-0.7) | 0.63 |
| | | 80/20 | 0.75 (0.85-0.63) | 0.77 |
| | | 70/30 | 0.62 (0.75-0.49) | 0.83 |
| | | 60/40 | 0.72 (0.82-0.61) | 0.83 |
| | | 50/50 | 0.64 (0.75-0.53) | 0.77 |
| | 600 | 90/10 | 0.86 (0.94-0.76) | 0.53 |
| | | 80/20 | 0.71 (0.83-0.59) | 0.69 |
| | | 70/30 | 0.62 (0.74-0.48) | 0.75 |
| | | 60/40 | 0.69 (0.78-0.57) | 0.8 |
| | | 50/50 | 0.73 (0.81-0.64) | 0.84 |
| | 800 | 90/10 | 0.86 (0.93-0.78) | 0.85 |
| | | 80/20 | 0.74 (0.83-0.64) | 0.73 |
| | | 70/30 | 0.62 (0.72-0.51) | 0.75 |
| | | 60/40 | 0.68 (0.77-0.59) | 0.84 |
| | | 50/50 | 0.68 (0.77-0.6) | 0.77 |
| | 1000 | 90/10 | 0.83 (0.9-0.75) | 0.7 |
| | | 80/20 | 0.71 (0.79-0.62) | 0.68 |
| | | 70/30 | 0.73 (0.82-0.64) | 0.79 |
| | | 60/40 | 0.7 (0.77-0.62) | 0.79 |
| | | 50/50 | 0.69 (0.77-0.6) | 0.76 |
| | 2000 | 90/10 | 0.87 (0.91-0.82) | 0.69 |
| | | 80/20 | 0.76 (0.81-0.69) | 0.76 |
| | | 70/30 | 0.67 (0.74-0.61) | 0.77 |
| | | 60/40 | 0.69 (0.75-0.63) | 0.83 |
| | | 50/50 | 0.69 (0.74-0.64) | 0.79 |
| | 3000 | 90/10 | 0.87 (0.91-0.84) | 0.76 |
| | | 80/20 | 0.77 (0.82-0.72) | 0.79 |
| | | 70/30 | 0.76 (0.81-0.71) | 0.85 |
| | | 60/40 | 0.74 (0.78-0.7) | 0.83 |
| | | 50/50 | 0.74 (0.77-0.7) | 0.82 |
| | 4000 | 90/10 | 0.9 (0.93-0.87) | 0.74 |
| | | 80/20 | 0.76 (0.8-0.72) | 0.78 |
| | | 70/30 | 0.73 (0.77-0.69) | 0.82 |
| | | 60/40 | 0.68 (0.72-0.64) | 0.83 |
| | | 50/50 | 0.72 (0.75-0.68) | 0.82 |
| | 5000 | 90/10 | 0.88 (0.91-0.85) | 0.73 |
| | | 80/20 | 0.8 (0.84-0.77) | 0.8 |
| | | 70/30 | 0.72 (0.76-0.68) | 0.83 |
| | | 60/40 | 0.7 (0.73-0.66) | 0.85 |
| | | 50/50 | 0.76 (0.79-0.73) | 0.86 |
| Random Forest | 200 | 90/10 | 0.81 (0.95-0.69) | 0.63 |
| | | 80/20 | 0.74 (0.95-0.54) | 0.78 |
| | | 70/30 | 0.77 (0.93-0.59) | 0.89 |
| | | 60/40 | 0.48 (0.71-0.27) | 0.75 |
| | | 50/50 | 0.75 (0.88-0.6) | 0.87 |
| | 400 | 90/10 | 0.95 (1-0.88) | 0.87 |
| | | 80/20 | 0.84 (0.95-0.73) | 0.74 |
| | | 70/30 | 0.62 (0.75-0.48) | 0.76 |
| | | 60/40 | 0.71 (0.83-0.59) | 0.76 |
| | | 50/50 | 0.63 (0.76-0.49) | 0.76 |
| | 500 | 90/10 | 0.82 (0.91-0.71) | 0.79 |
| | | 80/20 | 0.75 (0.85-0.63) | 0.83 |
| | | 70/30 | 0.5 (0.63-0.38) | 0.76 |
| | | 60/40 | 0.72 (0.82-0.6) | 0.83 |
| | | 50/50 | 0.65 (0.76-0.55) | 0.77 |
| | 600 | 90/10 | 0.86 (0.94-0.77) | 0.77 |
| | | 80/20 | 0.68 (0.8-0.57) | 0.78 |
| | | 70/30 | 0.62 (0.74-0.5) | 0.69 |
| | | 60/40 | 0.61 (0.72-0.5) | 0.76 |
| | | 50/50 | 0.61 (0.71-0.51) | 0.81 |
| | 800 | 90/10 | 0.86 (0.93-0.78) | 0.82 |
| | | 80/20 | 0.76 (0.84-0.65) | 0.8 |
| | | 70/30 | 0.62 (0.72-0.51) | 0.76 |
| | | 60/40 | 0.65 (0.75-0.56) | 0.84 |
| | | 50/50 | 0.67 (0.76-0.58) | 0.77 |
| | 1000 | 90/10 | 0.84 (0.91-0.77) | 0.8 |
| | | 80/20 | 0.68 (0.77-0.6) | 0.78 |
| | | 70/30 | 0.7 (0.79-0.61) | 0.87 |
| | | 60/40 | 0.6 (0.69-0.5) | 0.81 |
| | | 50/50 | 0.6 (0.69-0.52) | 0.7 |
| | 2000 | 90/10 | 0.87 (0.91-0.82) | 0.7 |
| | | 80/20 | 0.73 (0.8-0.67) | 0.76 |
| | | 70/30 | 0.65 (0.71-0.59) | 0.8 |
| | | 60/40 | 0.61 (0.66-0.54) | 0.75 |
| | | 50/50 | 0.65 (0.7-0.6) | 0.74 |
| | 3000 | 90/10 | 0.88 (0.91-0.84) | 0.76 |
| | | 80/20 | 0.75 (0.79-0.7) | 0.79 |
| | | 70/30 | 0.67 (0.72-0.61) | 0.8 |
| | | 60/40 | 0.66 (0.71-0.61) | 0.78 |
| | | 50/50 | 0.72 (0.76-0.67) | 0.81 |
| | 4000 | 90/10 | 0.88 (0.91-0.85) | 0.76 |
| | | 80/20 | 0.74 (0.78-0.69) | 0.76 |
| | | 70/30 | 0.67 (0.72-0.63) | 0.81 |
| | | 60/40 | 0.6 (0.64-0.56) | 0.79 |
| | | 50/50 | 0.67 (0.71-0.63) | 0.77 |
| | 5000 | 90/10 | 0.86 (0.89-0.83) | 0.84 |
| | | 80/20 | 0.74 (0.79-0.71) | 0.8 |
| | | 70/30 | 0.64 (0.68-0.6) | 0.81 |
| | | 60/40 | 0.64 (0.69-0.6) | 0.79 |
| | | 50/50 | 0.71 (0.75-0.68) | 0.81 |
| Stochiastic Gradient Descent | 200 | 90/10 | 0.81 (0.95-0.64) | 0.69 |
| | | 80/20 | 0.85 (0.95-0.7) | 0.82 |
| | | 70/30 | 0.77 (0.91-0.62) | 0.78 |
| | | 60/40 | 0.67 (0.84-0.49) | 0.76 |
| | | 50/50 | 0.69 (0.85-0.53) | 0.73 |
| | 400 | 90/10 | 0.97 (1-0.92) | 0.97 |
| | | 80/20 | 0.88 (0.95-0.8) | 0.9 |
| | | 70/30 | 0.67 (0.79-0.53) | 0.78 |
| | | 60/40 | 0.68 (0.79-0.56) | 0.71 |
| | | 50/50 | 0.72 (0.83-0.6) | 0.83 |
| | 500 | 90/10 | 0.88 (0.96-0.79) | 0.6 |
| | | 80/20 | 0.84 (0.92-0.75) | 0.74 |
| | | 70/30 | 0.78 (0.87-0.68) | 0.86 |
| | | 60/40 | 0.8 (0.89-0.7) | 0.81 |
| | | 50/50 | 0.74 (0.83-0.65) | 0.78 |
| | 600 | 90/10 | 0.91 (0.97-0.84) | 0.65 |
| | | 80/20 | 0.74 (0.84-0.63) | 0.64 |
| | | 70/30 | 0.72 (0.82-0.63) | 0.73 |
| | | 60/40 | 0.62 (0.72-0.52) | 0.73 |
| | | 50/50 | 0.72 (0.81-0.63) | 0.79 |
| | 800 | 90/10 | 0.9 (0.96-0.84) | 0.88 |
| | | 80/20 | 0.84 (0.9-0.75) | 0.67 |
| | | 70/30 | 0.78 (0.85-0.7) | 0.74 |
| | | 60/40 | 0.7 (0.78-0.62) | 0.75 |
| | | 50/50 | 0.68 (0.76-0.6) | 0.73 |
| | 1000 | 90/10 | 0.89 (0.94-0.84) | 0.78 |
| | | 80/20 | 0.75 (0.82-0.68) | 0.7 |
| | | 70/30 | 0.78 (0.84-0.7) | 0.81 |
| | | 60/40 | 0.78 (0.85-0.71) | 0.81 |
| | | 50/50 | 0.72 (0.79-0.65) | 0.82 |
| | 2000 | 90/10 | 0.89 (0.92-0.84) | 0.69 |
| | | 80/20 | 0.87 (0.91-0.83) | 0.83 |
| | | 70/30 | 0.79 (0.84-0.75) | 0.84 |
| | | 60/40 | 0.8 (0.84-0.76) | 0.86 |
| | | 50/50 | 0.78 (0.82-0.73) | 0.86 |
| | 3000 | 90/10 | 0.92 (0.95-0.9) | 0.85 |
| | | 80/20 | 0.87 (0.9-0.84) | 0.82 |
| | | 70/30 | 0.81 (0.85-0.78) | 0.85 |
| | | 60/40 | 0.81 (0.84-0.77) | 0.88 |
| | | 50/50 | 0.83 (0.86-0.8) | 0.87 |
| | 4000 | 90/10 | 0.93 (0.95-0.9) | 0.83 |
| | | 80/20 | 0.88 (0.9-0.85) | 0.89 |
| | | 70/30 | 0.86 (0.88-0.83) | 0.9 |
| | | 60/40 | 0.82 (0.85-0.79) | 0.89 |
| | | 50/50 | 0.8 (0.83-0.76) | 0.88 |
| | 5000 | 90/10 | 0.9 (0.93-0.88) | 0.85 |
| | | 80/20 | 0.88 (0.9-0.85) | 0.87 |
| | | 70/30 | 0.84 (0.87-0.81) | 0.89 |
| | | 60/40 | 0.82 (0.85-0.79) | 0.9 |
| | | 50/50 | 0.84 (0.87-0.82) | 0.9 |
| Support Vector Classifier | 200 | 90/10 | 0.82 (0.98-0.64) | 0.65 |
| | | 80/20 | 0.79 (0.93-0.6) | 0.78 |
| | | 70/30 | 0.77 (0.93-0.59) | 0.82 |
| | | 60/40 | 0.37 (0.57-0.18) | 0.84 |
| | | 50/50 | 0.55 (0.72-0.36) | 0.79 |
| | 400 | 90/10 | 0.95 (1-0.88) | 0.95 |
| | | 80/20 | 0.84 (0.94-0.72) | 0.9 |
| | | 70/30 | 0.69 (0.82-0.54) | 0.79 |
| | | 60/40 | 0.74 (0.86-0.61) | 0.76 |
| | | 50/50 | 0.61 (0.73-0.48) | 0.84 |
| | 500 | 90/10 | 0.82 (0.91-0.71) | 0.7 |
| | | 80/20 | 0.77 (0.88-0.65) | 0.8 |
| | | 70/30 | 0.68 (0.8-0.56) | 0.85 |
| | | 60/40 | 0.74 (0.83-0.63) | 0.83 |
| | | 50/50 | 0.71 (0.81-0.61) | 0.79 |
| | 600 | 90/10 | 0.86 (0.94-0.77) | 0.64 |
| | | 80/20 | 0.73 (0.85-0.61) | 0.76 |
| | | 70/30 | 0.71 (0.8-0.6) | 0.78 |
| | | 60/40 | 0.7 (0.8-0.6) | 0.77 |
| | | 50/50 | 0.73 (0.82-0.65) | 0.82 |
| | 800 | 90/10 | 0.86 (0.93-0.78) | 0.89 |
| | | 80/20 | 0.81 (0.89-0.72) | 0.77 |
| | | 70/30 | 0.75 (0.83-0.66) | 0.77 |
| | | 60/40 | 0.73 (0.81-0.66) | 0.79 |
| | | 50/50 | 0.71 (0.78-0.63) | 0.76 |
| | 1000 | 90/10 | 0.84 (0.91-0.77) | 0.8 |
| | | 80/20 | 0.74 (0.82-0.66) | 0.74 |
| | | 70/30 | 0.81 (0.87-0.74) | 0.8 |
| | | 60/40 | 0.75 (0.82-0.68) | 0.84 |
| | | 50/50 | 0.75 (0.81-0.68) | 0.83 |
| | 2000 | 90/10 | 0.89 (0.93-0.84) | 0.71 |
| | | 80/20 | 0.82 (0.87-0.77) | 0.83 |
| | | 70/30 | 0.82 (0.87-0.77) | 0.86 |
| | | 60/40 | 0.83 (0.87-0.78) | 0.89 |
| | | 50/50 | 0.78 (0.82-0.72) | 0.86 |
| | 3000 | 90/10 | 0.93 (0.95-0.9) | 0.83 |
| | | 80/20 | 0.86 (0.89-0.82) | 0.83 |
| | | 70/30 | 0.84 (0.88-0.8) | 0.86 |
| | | 60/40 | 0.84 (0.87-0.81) | 0.9 |
| | | 50/50 | 0.83 (0.86-0.79) | 0.9 |
| | 4000 | 90/10 | 0.92 (0.94-0.89) | 0.85 |
| | | 80/20 | 0.86 (0.89-0.82) | 0.87 |
| | | 70/30 | 0.86 (0.89-0.83) | 0.9 |
| | | 60/40 | 0.86 (0.89-0.84) | 0.92 |
| | | 50/50 | 0.83 (0.85-0.8) | 0.91 |
| | 5000 | 90/10 | 0.92 (0.94-0.89) | 0.82 |
| | | 80/20 | 0.86 (0.89-0.84) | 0.85 |
| | | 70/30 | 0.87 (0.89-0.85) | 0.91 |
| | | 60/40 | 0.86 (0.88-0.84) | 0.92 |
| | | 50/50 | 0.86 (0.89-0.84) | 0.93 |
| BERT | 200 | 90/10 | 0.85 (0.71-0.96) | 0.50 |
| | | 80/20 | 0.63 (0.45-0.78) | 0.50 |
| | | 70/30 | 0.45 (0.28-0.61) | 0.50 |
| | | 60/40 | 0.61 (0.44-0.76) | 0.59 |
| | | 50/50 | 0.6 (0.47-0.75) | 0.60 |
| | 400 | 90/10 | 0.82 (0.71-0.91) | 0.50 |
| | | 80/20 | 0.61 (0.5-0.73) | 0.50 |
| | | 70/30 | 0.76 (0.65-0.86) | 0.64 |
| | | 60/40 | 0.68 (0.58-0.78) | 0.67 |
| | | 50/50 | 0.54 (0.41-0.67) | 0.58 |
| | 500 | 90/10 | 0.72 (0.62-0.83) | 0.50 |
| | | 80/20 | 0.79 (0.69-0.88) | 0.59 |
| | | 70/30 | 0.57 (0.46-0.69) | 0.52 |
| | | 60/40 | 0.62 (0.52-0.71) | 0.62 |
| | | 50/50 | 0.65 (0.56-0.74) | 0.67 |
| | 600 | 90/10 | 0.81 (0.73-0.89) | 0.50 |
| | | 80/20 | 0.65 (0.54-0.77) | 0.52 |
| | | 70/30 | 0.51 (0.4-0.62) | 0.51 |
| | | 60/40 | 0.67 (0.58-0.76) | 0.67 |
| | | 50/50 | 0.7 (0.63-0.78) | 0.70 |
| | 800 | 90/10 | 0.82 (0.75-0.89) | 0.50 |
| | | 80/20 | 0.69 (0.61-0.79) | 0.57 |
| | | 70/30 | 0.77 (0.71-0.84) | 0.69 |
| | | 60/40 | 0.72 (0.65-0.79) | 0.72 |
| | | 50/50 | 0.72 (0.65-0.8) | 0.73 |
| | 1000 | 90/10 | 0.86 (0.8-0.91) | 0.50 |
| | | 80/20 | 0.74 (0.66-0.81) | 0.50 |
| | | 70/30 | 0.79 (0.74-0.85) | 0.73 |
| | | 60/40 | 0.74 (0.68-0.8) | 0.73 |
| | | 50/50 | 0.73 (0.67-0.78) | 0.72 |
| | 2000 | 90/10 | 0.87 (0.83-0.91) | 0.59 |
| | | 80/20 | 0.83 (0.79-0.87) | 0.73 |
| | | 70/30 | 0.78 (0.74-0.82) | 0.72 |
| | | 60/40 | 0.75 (0.7-0.79) | 0.74 |
| | | 50/50 | 0.78 (0.73-0.82) | 0.78 |
| | 3000 | 90/10 | 0.89 (0.85-0.91) | 0.58 |
| | | 80/20 | 0.83 (0.8-0.87) | 0.70 |
| | | 70/30 | 0.8 (0.76-0.83) | 0.75 |
| | | 60/40 | 0.77 (0.73-0.8) | 0.76 |
| | | 50/50 | 0.76 (0.73-0.8) | 0.76 |
| | 4000 | 90/10 | 0.88 (0.85-0.9) | 0.60 |
| | | 80/20 | 0.84 (0.82-0.87) | 0.73 |
| | | 70/30 | 0.81 (0.78-0.84) | 0.79 |
| | | 60/40 | 0.81 (0.78-0.83) | 0.79 |
| | | 50/50 | 0.79 (0.76-0.82) | 0.80 |
| | 5000 | 90/10 | 0.91 (0.89-0.93) | 0.68 |
| | | 80/20 | 0.85 (0.82-0.87) | 0.72 |
| | | 70/30 | 0.82 (0.79-0.84) | 0.79 |
| | | 60/40 | 0.8 (0.78-0.83) | 0.80 |
| | | 50/50 | 0.83 (0.8-0.85) | 0.83 |